\documentclass[10pt,journal,compsoc]{IEEEtran}
\ifCLASSOPTIONcompsoc
  \usepackage[nocompress]{cite}
\else
  \usepackage{cite}
\fi
\ifCLASSINFOpdf
\else
\fi

\usepackage{times}
\usepackage{epsfig}
\usepackage{graphicx}
\usepackage{amsmath}
\usepackage{amssymb}
\usepackage{multirow}
\usepackage{dutchcal}
\usepackage{xcolor}
\usepackage{subcaption}
\usepackage{colortbl}
\usepackage{caption}

\begin{document}
%
\title{
Self-Supervised 3D Action Representation Learning with Skeleton Cloud Colorization 
}
%
%
%
%

\author{Siyuan Yang,
        Jun Liu,
        Shijian Lu,
        Er Meng Hwa,~\IEEEmembership{Life~Fellow,~IEEE},\\
        Yongjian Hu,
        and~Alex C. Kot,~\IEEEmembership{Life~Fellow,~IEEE}
\IEEEcompsocitemizethanks{
\IEEEcompsocthanksitem Siyuan Yang is with the Rapid-Rich Object Search Lab, Interdisciplinary Graduate Programme, Nanyang Technological University, Singapore.\protect\\
E-mail: SIYUAN005@e.ntu.edu.sg
\IEEEcompsocthanksitem Jun Liu is with the Information Systems Technology and Design Pillar, Singapore University of Technology and Design, Singapore.\protect\\
E-mail: jun\_liu@sutd.edu.sg 
\IEEEcompsocthanksitem Shijian Lu is with the School of Computer Science \& Engineering, Nanyang Technological University, Singapore.\protect\\
E-mail: Shijian.Lu@ntu.edu.sg 
\IEEEcompsocthanksitem Er Meng Hwa and Alex C. Kot are with the School of Electrical and Electronic Engineering, Nanyang Technological University, Singapore.\protect\\
E-mail: \{emher, eackot\}@ntu.edu.sg
\IEEEcompsocthanksitem Yongjian Hu is with the South China University of Technology, Guangzhou, China.\protect\\
E-mail: eeyjhu@scut.edu.cn
}
\thanks{Corresponding author: Jun Liu}
}

%
%

\markboth{Journal of \LaTeX\ Class Files,~Vol.~14, No.~8, August~2015}%
{Shell \MakeLowercase{\textit{et al.}}: Bare Demo of IEEEtran.cls for Computer Society Journals}
%



\IEEEtitleabstractindextext{
\begin{abstract}
3D Skeleton-based human action recognition has attracted increasing attention in recent years. Most of the existing work focuses on supervised learning which requires a large number of labeled action sequences that are often expensive and time-consuming to annotate. 
In this paper, we address self-supervised 3D action representation learning for skeleton-based action recognition.
We investigate self-supervised representation learning
and design a novel skeleton cloud colorization technique that is capable of learning spatial and temporal skeleton representations from unlabeled skeleton sequence data. We represent a skeleton action sequence as a 3D skeleton cloud and colorize each point in the cloud according to its temporal and spatial orders in the original (unannotated) skeleton sequence. Leveraging the colorized skeleton point cloud, we design an auto-encoder framework that can learn spatial-temporal features from the artificial color labels of skeleton joints effectively. 
Specifically, we design a two-steam pretraining network that leverages fine-grained and coarse-grained colorization to learn multi-scale spatial-temporal features.
In addition, we design a Masked Skeleton Cloud Repainting task that can pretrain the designed auto-encoder framework to learn informative representations.
We evaluate our skeleton cloud colorization approach with linear classifiers trained under different configurations, including unsupervised, semi-supervised, fully-supervised, and transfer learning settings.
Extensive experiments on NTU RGB+D, NTU RGB+D 120, PKU-MMD, NW-UCLA, and UWA3D datasets show that the proposed method outperforms existing unsupervised and semi-supervised 3D action recognition methods by large margins and achieves competitive performance in supervised 3D action recognition as well.
\end{abstract}

}

\maketitle

\begin{figure*}[t]
\begin{center}
\includegraphics[trim=0cm 0.1cm 0.2cm 0cm,clip,width=0.9\textwidth]{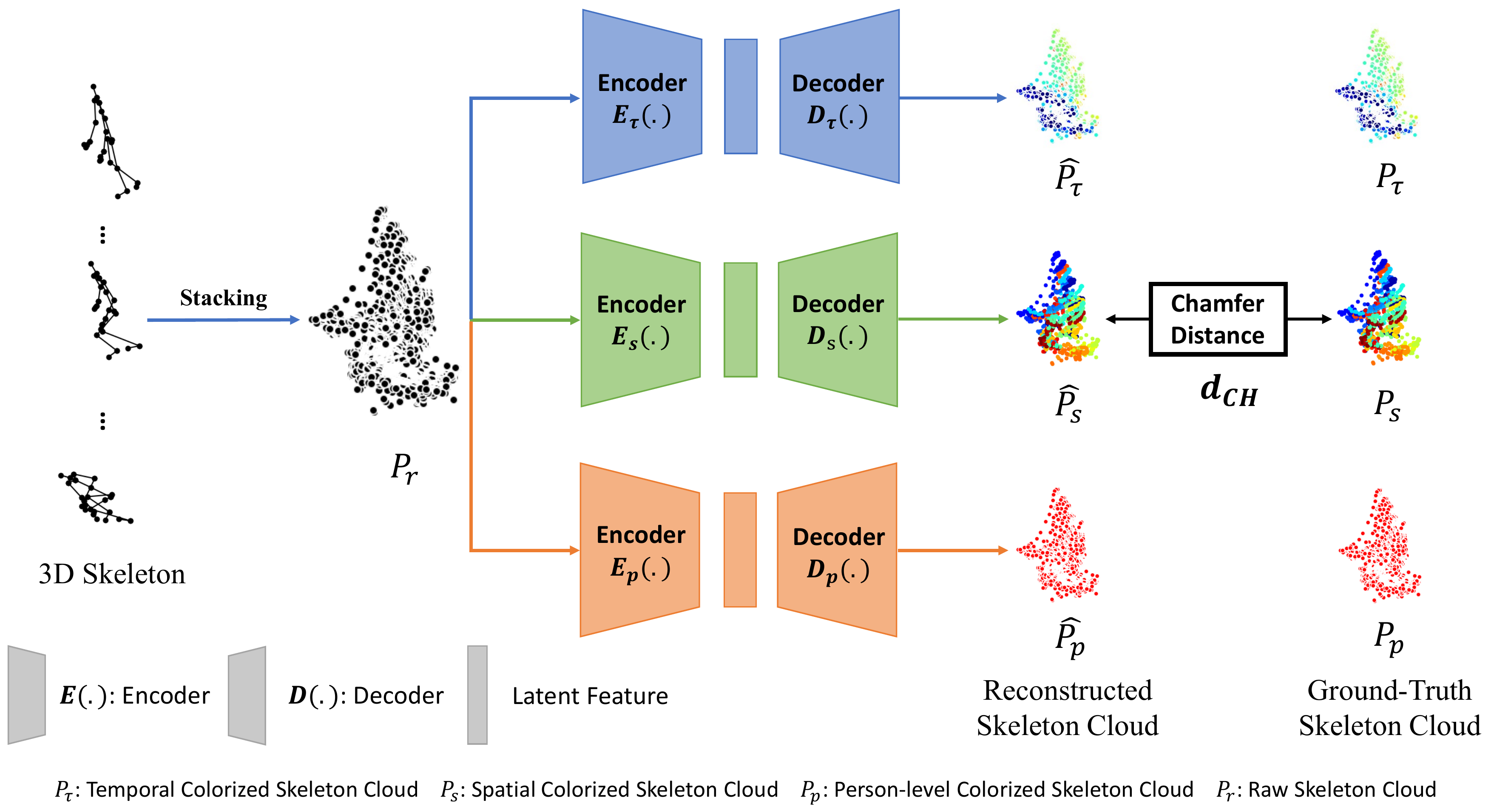}
\end{center}
\caption{
The pipeline of our proposed self-supervised representation learning
with skeleton cloud colorization.
Given a 3D skeleton sequence, we first stack it into a raw skeleton cloud $P_{r}$ and then colorize it into 3 skeleton clouds $P_{\tau}$, $P_{s}$, and $P_{p}$ (construction details shown in Fig. \ref{fig:color_pipeline_temporal} (a), Fig. \ref{fig:color_pipeline_spatial} (a) and Fig.~\ref{fig:person pipeline}) according to spatial, temporal, and person-level information, respectively. With the three colorized clouds as self-supervision signals, three encoder-decoders (with the same structure but no weight sharing) learn discriminative skeleton representative features.
(the encoder and decode details are provided in the Appendix)
}
\label{fig:framework}
\end{figure*}

\IEEEdisplaynontitleabstractindextext

%
\IEEEpeerreviewmaketitle


%
%
%
%

\IEEEraisesectionheading{\section{Introduction}\label{sec:introduction}}
\IEEEPARstart{H}uman action recognition is a fast-developing area due to its relevance to a wide range of applications in human-computer interaction, video surveillance, game control, etc. 
According to the types of input data, human action recognition can be grouped into different categories such as RGB-based \cite{carreira2017quo, simonyan2014two, wang2018temporal}, depth-based \cite{ oreifej2013hon4d, rahmani2014hopc, wang20203dv}, and 3D skeleton-based \cite{ke2017new, liu2017skeleton, liu2016spatio, Shi_2019_CVPR_twostream}, etc. 
Among these types of data modalities, 3D skeleton sequences, which represent a human body by the locations of keypoints in the 3D space and characterize informative human motions, have attracted increasing attention in recent years.
Compared with RGB videos or Depth videos, 3D skeleton data encodes high-level representations of human behaviors and is generally lightweight and robust to variations in appearances, surrounding distractions, viewpoint changes, etc. 
Additionally, with the development of depth sensors (\textit{\textit{e.g.,}} Microsoft Kinect, Asus Xtion, and Intel RealSense3D), skeleton sequences can be easily captured 
which triggers a large number of supervised methods that have been designed to learn spatio-temporal representations for skeleton-based action recognition.

Specifically, deep neural networks have been widely studied to model the spatio-temporal representation of skeleton sequences under supervised scenarios \cite{ke2017new, liu2017skeleton, liu2016spatio, Shi_2019_CVPR_twostream}. 
For example, Recurrent Neural Networks (RNNs) have been explored for modeling skeleton actions since they can capture temporal relations well \cite{du2015hierarchical, liu2017skeleton,liu2016spatio,Shahroudy_2016_CVPR, zhang2017view}. 
Convolutional Neural Networks (CNNs) have also been explored to build skeleton-based recognition frameworks by converting joint coordinates to 2D pseudo-images \cite{du2015skeleton, ke2017new, li2017skeleton, wang2017scene}. 
Recently, graph convolutional networks (GCNs), which generalize convolutional neural networks (CNNs) to graphs structures, have achieved increasing attention and have been adopted in many studies
\cite{bruna2013spectral, defferrard2016convolutional, welling2016semi, duvenaud2015convolutional, hamilton2017inductive, kipf2018neural} 
with outstanding performance \cite{peng2020learning, Shi_2019_CVPR_twostream, yan2018spatial}. 
However, most of these methods are fully-supervised which require a large number of labeled training samples that are often costly and time-consuming to collect. 
How to learn effective feature representations with minimal annotations becomes critically important. 
Recently,
several work \cite{kundu2019unsupervised, lin2020ms2l, su2020predict, zheng2018unsupervised, nie2020view, Li_2021_CVPR}
explores representation learning from unlabeled skeleton data for the task of skeleton action recognition, and the main technical line is to reconstruct skeleton data from the encoded features via certain encoder-decoder structures.
{Though remarkable progress has been achieved, self-supervised skeleton-based action representation learning remains a concerned problem.}

In this work, we propose to represent a skeleton sequence as a 3D skeleton cloud, and design a self-supervised representation learning scheme that learns features from spatial and temporal color labels. We treat a skeletal sequence as a spatial-temporal skeleton cloud by stacking the skeleton data of all frames together and colorizing each point (\textbf{fine-grained colorization}) in the cloud according to its temporal and spatial orders in the original skeleton sequence. Specifically, we learn spatial-temporal features from the corresponding joints' colors by leveraging a point-cloud based auto-encoder framework as shown in Fig.~\ref{fig:framework}. 
By repainting the whole skeleton cloud, our network can achieve self-supervised skeleton representation learning successfully by learning both spatial and temporal information from skeleton sequences.

The above-mentioned colorization manners only focus on learning the spatial and temporal information in the single-frame and single-joint levels, ignoring the temporal dependence across frames and the spatial relation between different pairs of joints. 
To tackle this issue, we propose a new type of colorization manner (\textbf{coarse-grained colorization}), which colorizes each point in the cloud according to the order of its corresponding multi-frame segment and body parts.
On top of that, we design a two-stream auto-encoder framework that learns the skeleton representation from both fine-grained and coarse-grained color labels and aligns the learned representations between different levels of spatial and temporal colorization.

Inspired by the Mask Auto-Encoder (MAE), we also design a Masked Skeleton Cloud Repainting task to pretrain the designed auto-encoder framework, 
aiming for learning more discriminative and informative self-supervised representations.
Specific to the skeleton action recognition task, we design five mask sampling strategies: random masking, temporal-only masking, segment masking, spatial-only masking, and body-part masking.

The contributions of this paper are summarized as follows:
\begin{itemize}
    \item We formulate self-supervised action representation learning as a 3D skeleton cloud repainting problem, where each skeleton sequence is treated as a skeleton cloud and can be directly processed with a point cloud auto-encoder framework. 
    \item We propose a novel skeleton cloud colorization scheme that colorizes each point in the skeleton cloud based on its temporal and spatial orders in the skeleton sequence. The color labels `fabricate’ self-supervision signals, which boost self-supervised skeleton action representation learning significantly.
    \item We further extend the design of our skeleton colorization methods with the Masked Skeleton Cloud Repainting task and propose a more powerful coarse-fine alignment framework for better feature pre-training.  
    \item Extensive experiments show that our method outperforms state-of-the-art unsupervised and semi-supervised skeleton action recognition methods by large margins, and its performance is also on par with supervised skeleton-based action recognition methods.
\end{itemize}

This work is an extension of our preliminary conference paper \cite{Yang_2021_ICCV}. 
The new contributions of this work can be summarized in three major aspects.
In \cite{Yang_2021_ICCV}, 
\textit{First}, 
the colorization procedure in \cite{Yang_2021_ICCV} only considers self-supervision signals at single-frame and single-joint levels.
Such a supervision signal ignores temporal dependence across frames and the spatial relation between different pairs of joints.
In this work, we propose a new type of colorization manner (\textbf{coarse-grained colorization}) that colorizes each point according to the temporal order of the multi-frame segment and the spatial order of its corresponding body part. 
\textit{Second}, we design a novel two-stream auto-encoder network for learning both fine-grained and coarse-grained spatial-temporal information and aligning both scales' information to the main auto-encoder.
On top of that, we design a Mask Skeleton Cloud Repainting strategy to further improve the ability of self-supervised representation learning.
\textit{Third}, we extensively evaluate the proposed self-supervised representation learning framework on three more datasets, including the large-scale NTU RGBD 120 dataset, PKU-MMD dataset, and UWA3D dataset.
More extensive empirical analysis of the proposed approach is also provided in this paper.

The rest of this paper is organized as follows. We review the related works in Section \ref{Sec: related}. In Section \ref{sec: method}, we introduce our proposed masked self-supervised skeleton action representation learning method in detail. We present the experimental results, comparisons, and ablation studies in Section \ref{sec: experiment}. Finally, we conclude the paper in Section \ref{sec: conclusion}.

\section{Related work}
\label{Sec: related}

In this section, we first briefly review the supervised skeleton action recognition methods. 
We then introduce the self-supervised learning-based skeleton action recognition methods.

\subsection{Skeleton-based Action Recognition}
Recently, skeleton-based action recognition has attracted increasing interest in the research community.
Early works focused on extracting hand-craft features from skeleton sequences for action recognition. 
The hand-crafted feature-based methods can be roughly divided into joint-based and body part-based methods. 
Due to the strong feature learning capability, recent methods for skeleton-based action recognition pay more attention to deep networks. 
Deep-learning based skeleton action recognition methods employ Recurrent Neural Networks (RNNs), Convolutional Neural Networks (CNNs), and Graph Convolution Networks (GCNs) to learn skeleton-sequence representation directly. 
Specifically, RNNs have been widely used to model temporal dependencies and capture the motion features for skeleton-based action recognition. 
For example, \cite{du2015hierarchical} uses a hierarchical RNN model to represent human body structures and temporal dynamics of the body joints. \cite{liu2017skeleton,liu2016spatio} proposes a 2D Spatio-Temporal LSTM framework to employ the hidden sources of action-related information over both spatial and temporal domains concurrently. 
\cite{zhang2017view} adds a view-adaptation scheme to the LSTM to regulate the observation viewpoints.
\cite{song2017end} proposes a Deep LSTM with spatial-temporal attention, where a spatial attention sub-network and a temporal attention sub-network work jointly.

CNN-based methods~\cite{du2015skeleton, ke2017new, li2018co, li2017skeleton, soo2017interpretable, wang2017scene} have also been proposed for skeleton action recognition. They usually transform the skeleton sequences into skeleton maps of the same target size and then use CNNs to learn the spatial and temporal dynamics or apply temporal convolution on skeleton sequences. 
For example, \cite{du2015skeleton, li2017skeleton} transform a skeleton sequence to an image by treating the joint coordinate (x,y,z) as the R, G, and B channels of a pixel and then adopted CNNs for action recognition.
\cite{ke2017new} transforms the 3D skeleton sequence into three skeleton clips, which are then fed to a CNN network for robust action feature learning. 
\cite{wang2017scene} presents a “scene flow to action map” representation for action recognition with CNNs.
\cite{soo2017interpretable} utilizes the Temporal CNN (TCN), which provides a way to explicitly learn interpretable spatial-temporal representations for skeleton-based action recognition.
\cite{li2018co} designs an end-to-end convolutional framework for learning co-occurrence features with a  hierarchical methodology.

Inspired by the observation that the human 3D skeleton is naturally a topological graph, Graph Convolutional Networks (GCNs) have attracted increasing attention in skeleton-based action recognition. 
For example, \cite{yan2018spatial} presents a spatial-temporal GCN to learn both spatial and temporal patterns from skeleton data. 
\cite{Shi_2019_CVPR_twostream} uses a non-local method to learn the individual topology of graphs instead of using the manually designed one. Additionally, \cite{Shi_2019_CVPR_twostream} proposes a two-stream Adaptive GCN with the additional bone information.
\cite{peng2020learning} recognizes actions by searching for different graphs at different layers via neural architecture search.
To reduce computational costs of GCNs, \cite{cheng2020skeleton} designs a Shift-GCN, which leverages the shift graph operations and lightweight point-wise convolutions.
\cite{chen2021channel} proposes a Channel-wise Topology Refinement Graph Convolution (CTR-GC) to learn different topologies in different channels for skeleton-based action recognition.
More recently, \cite{chi2022infogcn} proposes InfoGCN, which combines an information bottleneck framework to learn informative representation and an attention-based graph convolution that infers context-dependent skeleton topology. 
Though the aforementioned methods achieve very impressive performance, they are all supervised, requiring a large amount of labeled data which is prohibitively time-consuming to collect. 
In this work, we study self-supervised representation learning in skeleton-based action recognition, which mitigates the data labeling constraint greatly.

\subsection{Self-Supervised Representation Learning for Skeleton Action Recognition}
{Self-Supervised action recognition aims to learn effective feature representations by predicting future frames of input sequences~\cite{lin2020ms2l, su2020predict}, by re-generating the sequences~\cite{zheng2018unsupervised, kundu2019unsupervised}, or by the contrastive pretext tasks~\cite{Li_2021_CVPR, guo2022contrastive, zhang2022hierarchical}.}
Most existing methods focus on RGB videos or RGB-D videos. {For example, \cite{srivastava2015unsupervised} uses an LSTM-based Encoder-Decoder architecture to learn video representations.} \cite{luo2017unsupervised} uses an RNN-based encoder-decoder framework to predict the sequences of flows computed with RGB-D modalities. \cite{li2018unsupervised} uses unlabeled video to learn view-invariant video representations.

\begin{figure*}[t]
\centering
\begin{minipage}{0.29\textwidth}
\begin{center}
\includegraphics[width=0.9\textwidth]{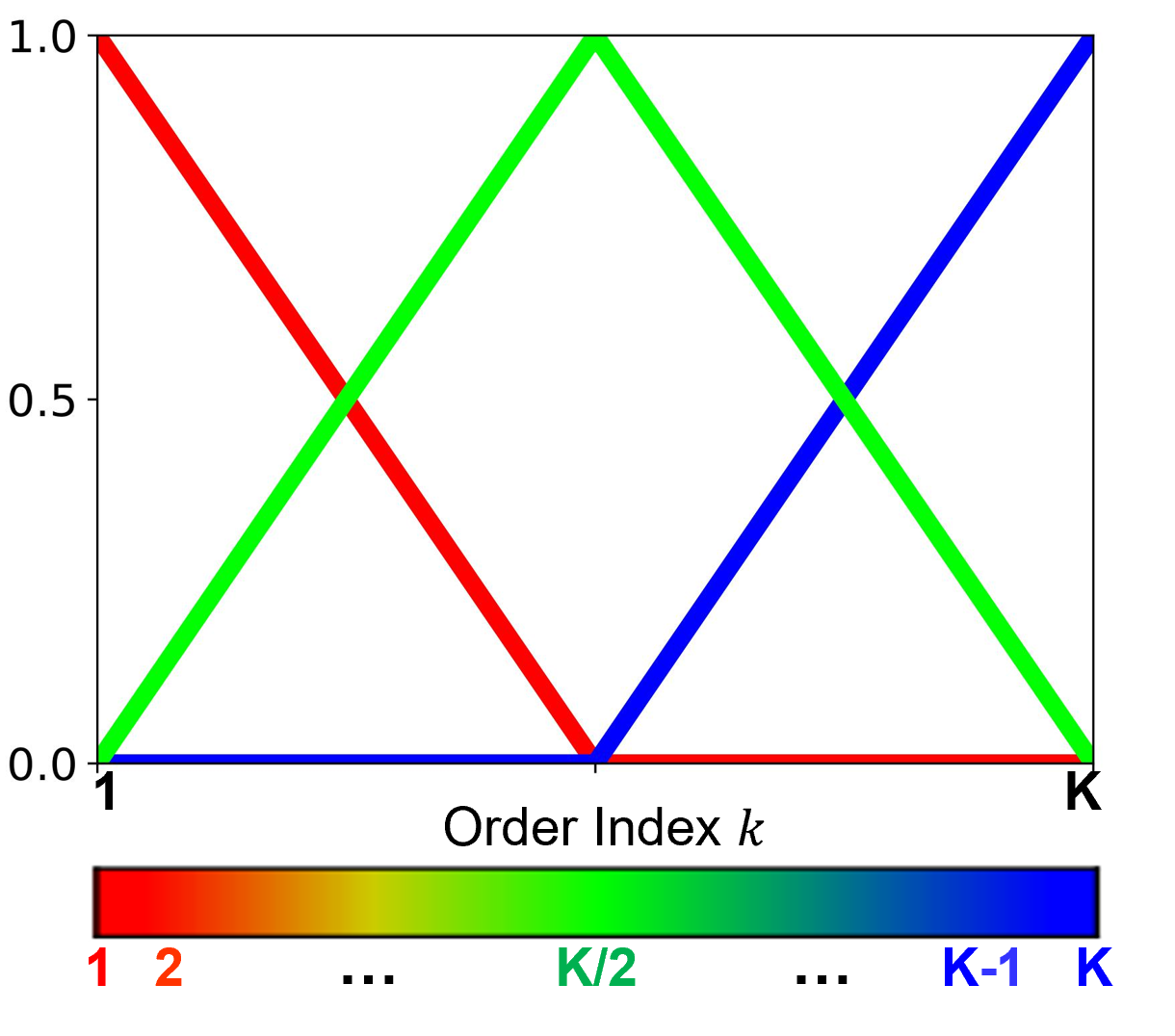}
\end{center}
\subcaption{Illustration of the colorization scheme. 
  With the increase of order index $k$ (where $k \in [1,K]$), the corresponding color changes from red to green to blue. (Best viewed in color) 
  }
\end{minipage}
\hfill
\begin{minipage}{0.32\textwidth}
\begin{equation}
\footnotesize 
r^{x}_{y, z} = \left\{
\begin{aligned}
-2\times (k/K) + 1 &, \;\mbox{if $k <= K/2$}\\
0  &, \;\mbox{if $k > K/2$}\\
\end{aligned}
\right.
\end{equation}
\begin{equation}
\footnotesize 
g^{x}_{y, z} = \left\{
\begin{aligned}
2\times (k/K) & ,\;\mbox{if $k <= K/2$}\\
-2\times (k/K) + 2  &, \;\mbox{if $k > K/2$}\\
\end{aligned}
\right.
\end{equation}
\begin{equation}
\footnotesize 
b^{x}_{y, z} = \left\{
\begin{aligned}
0 & ,\;\mbox{if $k <= K/2$}\\
2\times (k/K) - 1  & ,\;\mbox{if $k > K/2$}\\
\end{aligned}
\right.
\end{equation}

\subcaption{The general formula of the R, G, B color distribution of Fig.~\ref{fig:colorize} (a).}
\end{minipage}
\hfill
\begin{minipage}{0.36\textwidth}
\begin{center}
\resizebox{0.98\textwidth}{!}{
\begin{tabular}{|l|l|c|c|c|c|c|} 
  \hline 
   & Colorization & x  &  y   & z & k & K \\
  \hline
   (i) & Temporal ($\tau$)                & $\tau$  & t  & j & t  & T        \\
  \hline
   (ii) &Spatial  (s)                & s       &  t  & j & j  &  J      \\
  \hline
   (iii) & Coarse-Temporal (ct)          & ct  & s  & j & s & S \\
   \hline
   (iv) & Coarse-Spatial    (cs)       & cs  & t  & $\mathcal{p}$ & $\mathcal{p}$ & $\mathcal{P}$ \\
  
  \hline
\end{tabular}
}
\end{center}
\subcaption{The specific parameters ($x, y, z, k, K$) selection for the designed four types of colorization scheme. t: temporal index; T: total number of frame; j: joint index; J: total number of joint; s: segment index; S: total number of segment; $\mathcal{p}$: part index; $\mathcal{P}$: total number of body part.}
\end{minipage}
\caption{The overall definition of our designed skeleton cloud colorization schemes. \textbf{(a)} Illustration of the colorization scheme. \textbf{Top}: Definition of each color channel (RGB) when varying $k$ (where $k \in [1,K]$). 
\textbf{Bottom}: The corresponding color of index $k$.
\textbf{(b)} The general formula of the R, G, B color distribution of Fig.~\ref{fig:colorize} (a).
\textbf{(c)} The specific parameters selection for the designed four types of colorization scheme. The processings and visualizations of temporal, spatial, coarse-temporal, coarse-spatial colorization can be found in Fig.~\ref{fig:color_pipeline_temporal} and Fig.~\ref{fig:color_pipeline_spatial}.
}
\label{fig:colorize}
\end{figure*}

Self-Supervised skeleton-based action recognition was largely neglected though a few works have attempted to address this challenging task very recently. 
For example, \cite{zheng2018unsupervised} presents a GAN encoder-decoder to re-generate masked input sequences. 
\cite{kundu2019unsupervised} adopts a hierarchical fusion approach to improve human motion generation. 
\cite{su2020predict} presents a decoder-weakening strategy to drive the encoder to learn discriminative action features.
With the development of contrastive learning, there are a few self-supervised contrastive skeleton-based action recognition methods emerged in the past two years.
\cite{rao2021augmented} use the momentum encoder \cite{he2020momentum} for contrastive learning with single-stream skeleton sequence. 
While \cite{Li_2021_CVPR} proposes a cross-stream knowledge mining strategy to improve the performance with multi-types of skeleton sequences.
\cite{guo2022contrastive} introduces an extreme augmentation strategy to force the model to learn more general representation by providing harder contrastive pairs.
\cite{GL-Transformer} proposes GL-Transformer, which is able to effectively capture the global context and local dynamics of the sequence.
\cite{zhang2022contrastive} follows the SimSiam~\cite{chen2021exploring} structure and introduces a novel positive-enhanced learning strategy for unsupervised skeleton representation learning.
\cite{zhang2022hierarchical} proposes a new hierarchical contrastive learning framework, HiCLR, to take advantage of the strong augmentations.

The aforementioned methods process skeleton sequences frame by frame and extract temporal features from ordered sequences or leverage contrastive learning to process the self-supervised skeleton action recognition by contrastive pairs. 
We instead treat a skeleton sequence as a novel colored skeleton cloud by stacking human joints of each frame together. 
We design a novel skeleton colorization scheme and leverage the color information for self-supervised spatial-temporal representation learning.
Additionally, we introduce two types of colorization strategies (fine-grained and coarse-grained colorization) by assigning colors from different spatial levels and different temporal levels.

\section{Method}
\label{sec: method}
In this section, we present our masked skeleton cloud colorization representation learning method that converts the skeleton sequence to a skeleton cloud and colorizes each point in the cloud by its spatial-temporal properties. In particular, we present how to construct the skeleton cloud in Section~\ref{Data_process} and describe the colorization step in Section~\ref{Skeleton Coloring}. 
In Section~\ref{sec: C-F}, we introduce the coarse-fine skeleton cloud colorization.
We introduce the repainting pipeline and masking strategy in Section~\ref{Pipeline} and Section~\ref{sec: mask}, respectively. 
Finally, the training details are described in Section~\ref{training}.

\subsection{Data Processing}\label{Data_process}
Given a skeleton sequence under the global coordinate, the $j^{th}$ skeleton joint in the $t^{th}$ frame is denoted as $v_{t, j} = [x_{t, j}, y_{t, j}, z_{t, j}]$, $t\in (1, \cdots, T)$, $j\in (1, \cdots, J)$, where $T$ and $J$ denote the number of frames and body joints, respectively. Generally, skeleton data is defined as a sequence, and the set of joints in the $t^{th}$ frame are denoted as $V_{t} = \left\{v_{t,j}|j = 1, ..., J  \right\}$. We propose to treat all the joints in a skeleton sequence as a whole by stacking all frames' data together, and Fig.~\ref{fig:framework} illustrates the stacking framework. We name the stacked data as skeleton cloud and denote it by $P_{r} = \left\{v_{t, j} = [x_{t, j}, y_{t, j}, z_{t, j}]|t = 1, ..., T; j = 1, ..., J \right\}$. Therefore, the obtained 3D skeleton cloud consists of $N = T\times J$ 3D points in total. We use $P_{r}$ to denote the raw skeleton cloud so as to differentiate it from the colorized clouds to be described later.

\subsection{Skeleton Cloud Colorization}\label{Skeleton Coloring}
Points within our skeleton cloud are positioned with 3D coordinates (x, y, z), which is similar to a normal point cloud that consists of unordered points. The spatial relation and temporal dependency of skeleton cloud points are crucial in skeleton-based action recognition, but they are largely neglected in the aforementioned raw skeleton cloud data. We propose an innovative skeleton cloud colorization method to exploit the spatial relation and temporal dependency of skeleton cloud points for skeleton-based action recognition.

\noindent\textbf{Temporal Colorization.} 
Temporal information is critical in action recognition. To assign each point in the skeleton cloud a temporal feature, we colorize the skeleton cloud points according to their relative time order (from $1$ to $T$) in the original skeleton sequence. 
{Different colorization schemes have been reported, and here we adopt the colorization scheme that uses 3 RGB channels \cite{choutas2018potion}, as illustrated in Fig.~\ref{fig:colorize} (a). This colorization scheme works by linear mapping, which can assign similar colors to points of adjacent frames and help the learning of temporal order information. The general formulation of the R, G, B color distribution for Fig.~\ref{fig:colorize} (a) is shown in Fig.~\ref{fig:colorize} (b), and the specific parameter selection can be found in Fig.~\ref{fig:colorize} (c)(i).
Based on this, the formulation for temporal colorization is:}
\begin{equation}
r^{\tau}_{t, j} = \left\{
\begin{aligned}
-2\times (t/T) + 1 &, \;\mbox{if $t <= T/2$}\\
0  &, \;\mbox{if $t > T/2$}\\
\end{aligned}
\right.
\label{temporal r}
\end{equation}

\begin{equation}
g^{\tau}_{t, j} = \left\{
\begin{aligned}
2\times (t/T) & ,\;\mbox{if $t <= T/2$}\\
-2\times (t/T) + 2  &, \;\mbox{if $t > T/2$}\\
\end{aligned}
\right.
\label{temporal b}
\end{equation}

\begin{equation}
b^{\tau}_{t, j} = \left\{
\begin{aligned}
0 & ,\;\mbox{if $t <= T/2$}\\
2\times (t/T) - 1  & ,\;\mbox{if $t > T/2$}\\
\end{aligned}
\right.
\label{temporal g}
\end{equation}

\begin{figure*}[t]
\begin{center}
\includegraphics[trim=0cm 0cm 0cm 0cm,clip,width=0.98\textwidth]{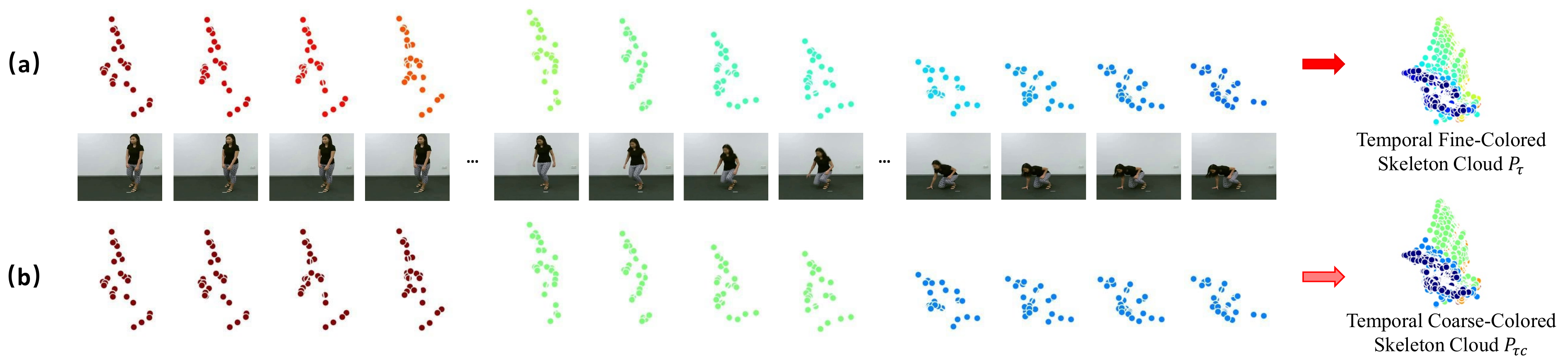}

\end{center}
\caption{
The pipelines of temporal colorization and temporal coarse-grained colorization. (a) Given a skeleton sequence, the temporal colorization colorizes points based on the relative temporal order $t$ ($t \in [1, T]$) in the sequential data. (b) The coarse-grained temporal colorization colorizes points based on the index of segments $s$ ($s \in [1, S]$). (Best viewed in color)  
}
\label{fig:color_pipeline_temporal}
\end{figure*}

With this colorizing scheme, we can assign different colors to points from different frames based on the frame index $t$ as illustrated in Fig.~\ref{fig:color_pipeline_temporal} (a). More specifically, with this temporal-index based colorization scheme, each point will have a 3-channels feature that can be visualized with red, green, and blue channels (RGB channels) to represent its temporal information. Together with the original 3D coordinate information, the temporally colorized skeleton cloud can be denoted by $P_{\tau} = \{v^{\tau}_{t, j} = [x_{t, j}, y_{t, j}, z_{t, j}, r^{\tau}_{t, j}, g^{\tau}_{t, j}, b^{\tau}_{t, j}]|t = 1, ..., T; j = 1, ..., J \}$.

\begin{figure}[t]
\begin{center}
\includegraphics[trim=0cm 0cm 0cm 0cm,clip,width=0.49\textwidth]{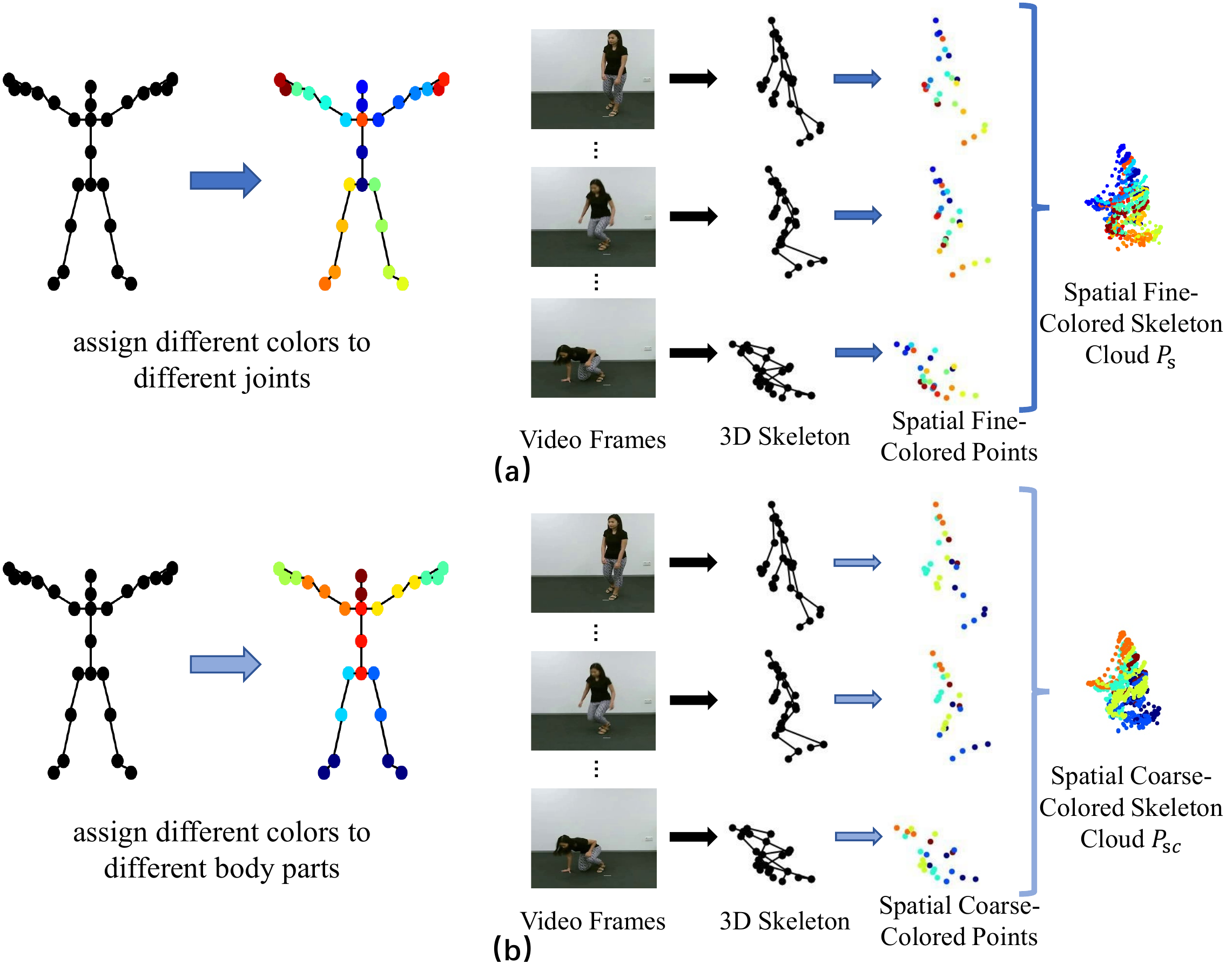}
\end{center}
\caption{
The pipelines of spatial colorization and spatial coarse-grained colorization. (a) Given a skeleton sequence, the spatial colorization colorizes points based on the relative spatial order $j$ ($j \in [1, J]$) in the sequential data. (b) The spatial colorization colorizes points based on the index of body part $\mathcal{p}$ ($\mathcal{p} \in [1, \mathcal{P}]$). (Best viewed in color)  
}
\label{fig:color_pipeline_spatial}
\end{figure}

\noindent\textbf{Spatial Colorization.} 
Besides temporal information, spatial information is also very important for action recognition. We employ a similar colorization scheme to colorize spatial information as illustrated in Fig.~\ref{fig:colorize} (a). The scheme assigns different colors to different points according to their spatial orders $j \in [1, J]$ ($J$ is the total number of joints in the skeleton cloud of a person), as shown in Fig. \ref{fig:color_pipeline_spatial} (a).
{The specific parameter selection for spatial colorization can be found in Fig.~\ref{fig:colorize} (c)(ii). 
The formulations are similar to Eq. \ref{temporal r}, \ref{temporal b}, \ref{temporal g}, while we replace the order index from temporal $T$ with spatial $J$, which can be found in the appendix.
}
We denote the spatially colorized skeleton cloud as $P_{s} = \{ v^{s}_{t, j} = [x_{t, j}, y_{t, j}, z_{t, j}, r^{s}_{t, j}, g^{s}_{t, j}, b^{s}_{t, j}]|t = 1, ..., T; j = 1, ..., J \}$. 
{With the increase of the spatial order index of the joint in the skeleton, points will be assigned with different colors that change from red to blue and to green gradually, which is able to represent the linear order information for the joint spatial order and facilitate the learning of spatial order information.}


\noindent\textbf{Person-level Colorization.} 
Human actions contain rich person interaction information as in NTU RGB+D \cite{Shahroudy_2016_CVPR}, which is important to the skeleton action recognition. We, therefore, propose a person-level colorization scheme for action recognition.

\begin{figure}[t]
\begin{center}
\includegraphics[trim=4cm 1.5cm 1.5cm 2cm,clip,width=0.4\textwidth]{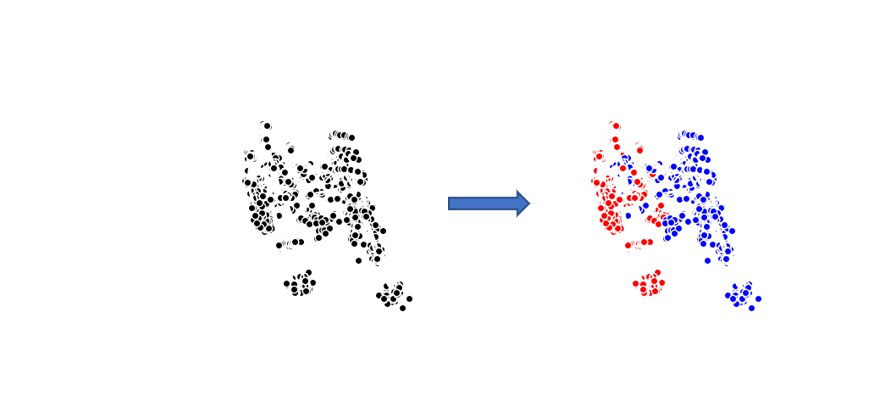}
\end{center}
  \caption{Person-level colorization. The first person's points will be assigned the red color, and person two will colorize to blue. 
  }
\label{fig:person pipeline}
\end{figure}

We focus on the scenarios that human interactions involve two persons only and apply different colors to the points of different persons. Specifically, we encode the first person's joints with red color and the second person's joints with blue color as illustrated in Fig.~\ref{fig:person pipeline}. The person-level colored clouds can thus be denoted by $P_{p} = \{v^{p}_{t, j, n} = [x_{t, j, n}, y_{t, j, n}, z_{t, j, n}, 1, 0, 0]| t = 1, ..., T; j = 1, ..., J; n = 1 \} \cup \{v^{p}_{t, j, n} = [x_{t, j, n}, y_{t, j, n}, z_{t, j, n}, 0, 0, 1]| t = 1, ..., T; j = 1, ..., J; n = 2 \} $, where $n = 1$ and $n = 2$ mean that the points belong to the first and the second persons, respectively. 

Given a raw skeleton cloud, the three colorization schemes thus construct three colorized skeleton clouds $P_{\tau}$, $P_{s}$ and $P_{p}$ that capture temporal dependency, spatial relations, and human interaction information, respectively.

\subsection{Coarse-Fine Skeleton Cloud Colorization}
\label{sec: C-F}
As mentioned above, in our framework, the skeleton cloud is colorized according to each point's temporal and spatial order information.  
Besides the frame-level and joint-level colorization (fine-grained colorization), coarse-grained colorization can also contribute to spatial-temporal feature learning for self-supervised skeleton action recognition.
This is because some actions are often performed at the body part level. 
For these actions, all the joints from the same informative body part tend to represent similar spatial information. 
Additionally, some actions contain long-range temporal dependence, which means that several consecutive frames contain similar representations.
These imply that coarse-grained skeleton cloud colorization is also useful for self-supervised skeleton representation learning.

\begin{figure*}[t]
\begin{minipage}{0.48\textwidth}
\begin{center}
\includegraphics[trim=0cm 0cm 0cm 0cm,clip,width=0.95\textwidth]{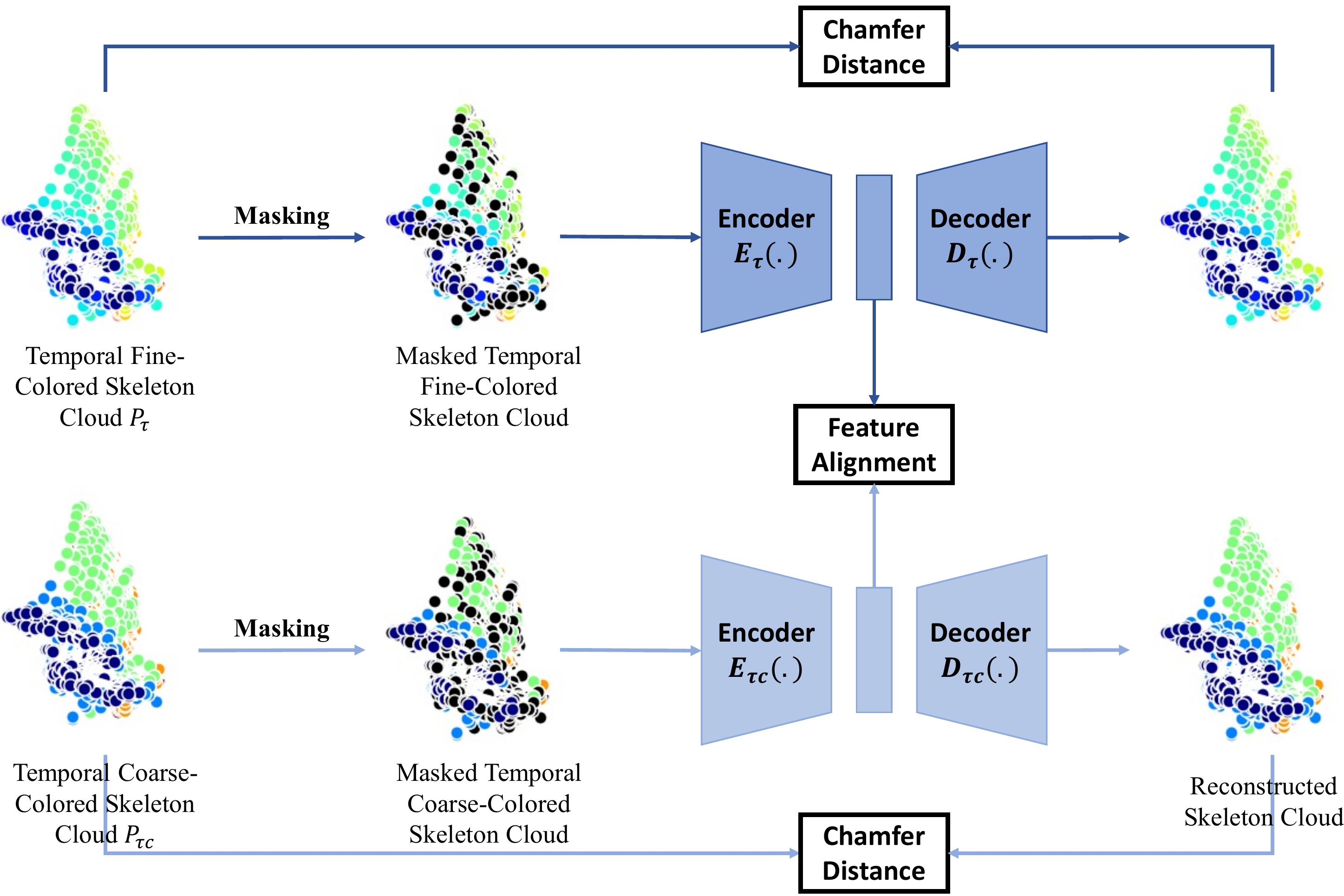}

\end{center}
\caption{
Illustration of the Coarse-Fine Alignment framework for temporal colorization, which incorporates both temporal fine-grained colorization and temporal coarse-grained colorization repainting. Utilizing the MSE loss, feature alignment is conducted to guide the main encoder (temporal encoder $E_{\tau}$) in capturing both the fine and coarse temporal information.    
}
\label{fig:align model temporal}
\end{minipage}
\hspace{4mm}
\begin{minipage}{0.48\textwidth}
\begin{center}
\includegraphics[trim=0cm 0cm 0cm 0cm,clip,width=0.95\textwidth]{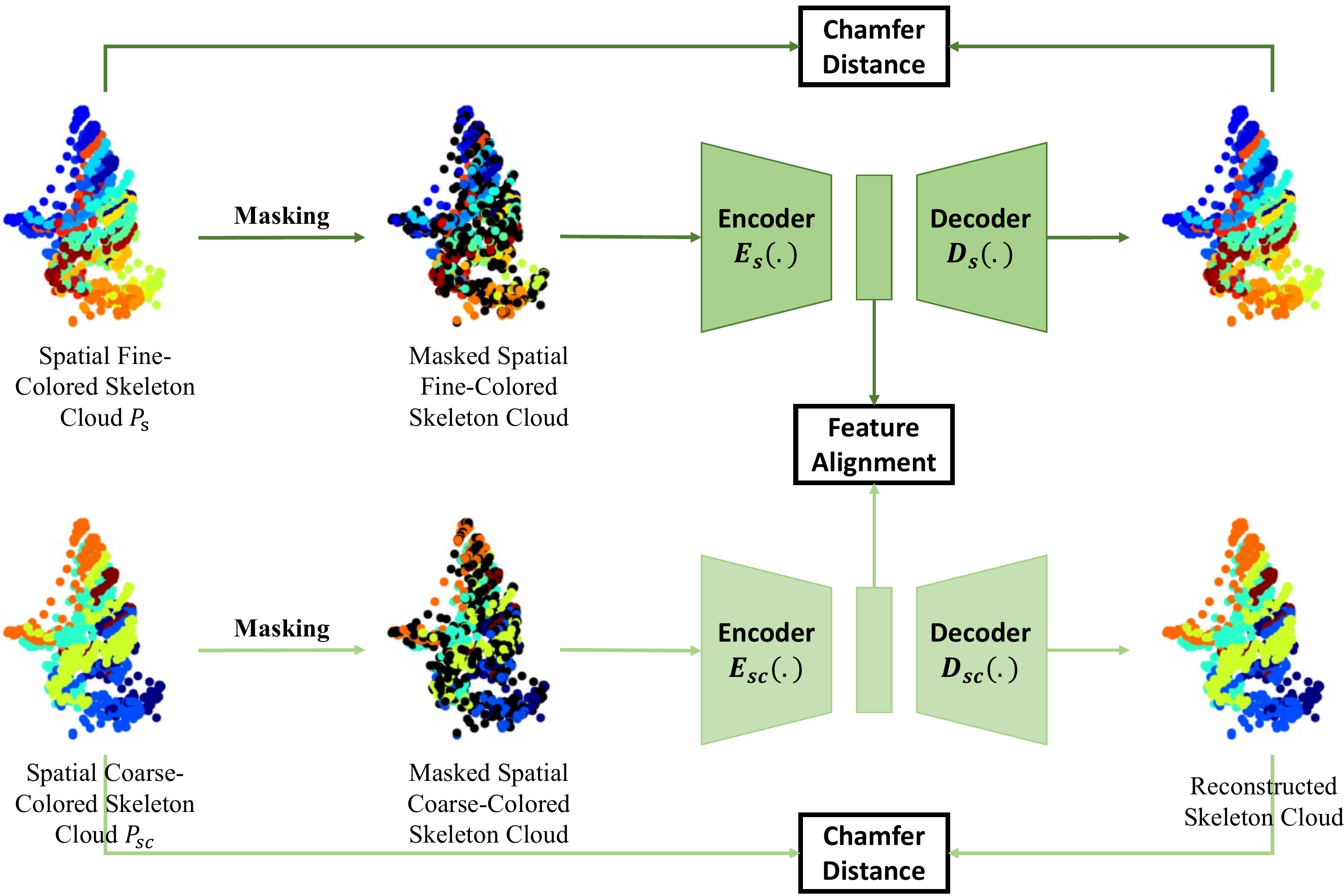}

\end{center}
\caption{
Illustration of the Coarse-Fine Alignment framework for spatial colorization, which contains both the fine-grained and coarse-grained spatial colorization pipelines. 
By implementing an alignment loss (MSE loss) at the latent feature level, we ensure that the latent feature of the fine-grained encoder $E_{s}$ captures both fine-grained and coarse-grained spatial information. 
}
\label{fig:align model spatial}
\end{minipage}
\end{figure*}

\noindent\textbf{Spatial Coarse-Grained Colorization.} 
Based on the human physical structure, the human skeleton can be segmented into multiple sections, comprising $\mathcal{P}$ parts, as outlined in \cite{du2015hierarchical,li2020dynamic}.
This allows us to undertake a coarse-grained spatial colorization aligned with the divisions of body parts.
In order to assign each point in the skeleton cloud a coarser spatial feature, we colorize the skeleton cloud points in accordance with the order of the body parts they correspond to (from 1 to $\mathcal{P}$). 
{The detailed parameter selection for the coarse-grained spatial colorization can be found in Fig.~\ref{fig:colorize} (c)(iii). 
The formulations are similar to Eq. \ref{temporal r}, \ref{temporal b}, \ref{temporal g}, while we replace the order index from the temporal $T$ order to the body part $\mathcal{P}$, which can be found in the appendix.}

{We use $P_{sc}$ to stand for the spatial coarse-grained colorized skeleton cloud. 
Different from the spatial colorization that assigns different colors to different joints (as shown in Fig.~\ref{fig:color_pipeline_spatial} (a)), the Spatial Coarse-Grained Colorization assigns different colors to points from different body parts, as shown in Fig. \ref{fig:color_pipeline_spatial} (b).}
In this way, the proposed spatial coarse-grained colorized skeleton cloud $P_{sc}$ contains the spatial relation information within different body parts.

\noindent\textbf{Temporal Coarse-Grained Colorization.}
Given that consecutive frames often possess similar representations, human skeleton sequences can be segmented into multiple sections, denoted as SS segments.
Here, we introduce a coarse-grained temporal colorization at the segment level. 
Specifically, we employ the colorization scheme (Fig.~\ref{fig:colorize} (a)) to colorize segment-level information. 
{For the coarse-grained temporal colorization. the detailed parameter selection are presented as in Fig.~\ref{fig:colorize} (c)(iv). 
The formulations are similar to Eq. \ref{temporal r}, \ref{temporal b}, \ref{temporal g}, while we substitute the order index from temporal $T$ with the segment $\mathcal{S}$. Further details are provided in the appendix.}

{Here, we use the $P_{\tau c}$ to stand for the temporal coarse-grained colorized skeleton cloud. 
Different from temporal colorization, which assigns different colors to individual frames (as shown in Fig.~\ref{fig:color_pipeline_temporal} (a)), Temporal Coarse-Grained Colorization assigns different colors to points from different segments, as shown in Fig.~\ref{fig:color_pipeline_temporal} (b).}
Consequently, the proposed temporal coarse-grained colorized skeleton cloud $P_{\tau c}$ encapsulates the temporal dependency information at the segment level.

Given a raw skeleton cloud, the two coarse-grained colorization schemes thus construct two colorized skeleton clouds $P_{\tau c}$ and $P_{sc}$ that capture temporal dependencies at the segment level and spatial relationships among distinct body parts, respectively.

\begin{figure}[t]
\begin{center}
\includegraphics[trim=0cm 0cm 0cm 0cm,clip,width=0.49\textwidth]{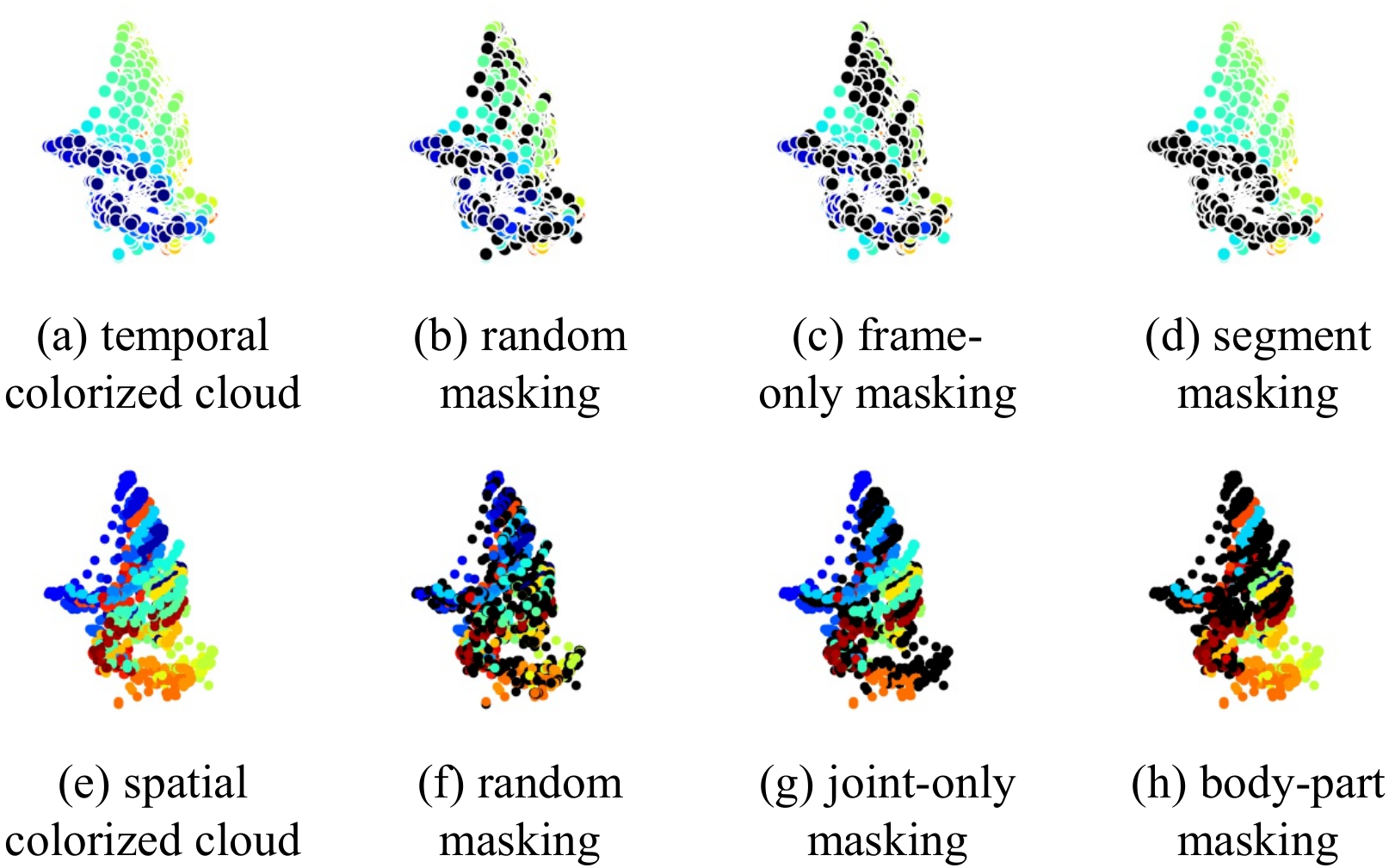}
\end{center}
  \caption{Mask Sampling Strategy. \textbf{(b)} and \textbf{(f)}: Random masking that is spacetime-agnostic. \textbf{(c)} Frame-only masking: mask all points from the randomly selected frames.
  \textbf{(d)} Segment masking: mask all points from the randomly selected continuous frames.
  \textbf{(g)} Joint-only masking: randomly mask points from selected joints, broadcasted to all frames.
  \textbf{(h)} Body-part masking: randomly mask points from selected body parts, broadcasted to all frames.
  (Best viewed in color)  
  }
\label{fig:masking}
\end{figure}

\subsection{Repainting Pipeline}\label{Pipeline}
Inspired by the success of self-supervised learning, our goal is to extract the temporal, spatial, and interactive information by learning to repaint the raw skeleton cloud $P_{r}$ in a self-supervised manner. As illustrated in Fig.~\ref{fig:framework}, we use colorized skeleton clouds (temporal-level $P_{\tau}$, spatial-level $P_{s}$, and person-level $P_{p}$) as three kinds of self-supervision signals, respectively. The framework consists of an encoder $E(.)$ and a decoder $D(.)$. Since we have three colorization schemes, we have three pairs of encoders ($E_{\tau}(.)$, $E_{s}(.)$, and $E_{p}(.)$) and decoders ($D_{\tau}(.)$, $D_{s}(.)$, and $D_{p}(.)$). Below we use the temporal colorization stream as an example to explain the model architecture and the training process.

\noindent\textbf{Model Architecture.}
As mentioned in Section \ref{Skeleton Coloring}, the obtained skeleton cloud format is similar to that of the normal point cloud. We therefore adopt DGCNN \cite{wang2019dynamic} (designed for point cloud classification and segmentation) as the backbone of our framework and use the modules before the fully-connected (FC) layers to build our encoder\footnote{Detailed network structure of encoder and decoder can be found in Appendix.\label{footnote}}.
In addition, we adopt the decoder\textsuperscript{\ref{footnote}} 
of FoldingNet \cite{yang2018foldingnet} as the decoder of our network architecture. Since the input and output of FoldingNet are all $N\times3$ matrices with 3D positions $(x, y, z)$ only, we enlarge the feature dimension to 6 to repaint both the position and color information. Assuming that the input is the raw point set $P_{r}$ and the obtained repainted point set is $\widehat{P_{\tau}} = D_{\tau}(E_{\tau}(P_{r}))$, the repainting error between the ground truth temporal colorization $P_{\tau}$ and the repainted $\widehat{P_{\tau}}$ is computed by using the Chamfer distance:
\begin{equation}
    d_{CH}(P_{\tau}, \widehat{P_{\tau}}) = Max \left\{A, B\right\},  where
\label{max}
\end{equation}

\begin{equation}
    A = \frac{1}{\left |P_{\tau} \right|} \sum_{v_{\tau}\in P_{\tau}} \mathop{\min} \limits_{\widehat{v_{\tau}}\in \widehat{P_{\tau}}} \left \|v_{\tau} - \widehat{v_{\tau}}\right \|_{2}
\end{equation}

\begin{equation}
    B = \frac{1}{\left |\widehat{P_{\tau}} \right|} \sum_{\widehat{v_{\tau}} \in \widehat{P_{\tau}}} \mathop{\min} \limits_{v_{\tau}\in P_{\tau}} \left \| \widehat{v_{\tau}} - v_{\tau} \right \|_{2}
\end{equation}

\noindent where the term $ \mathop{\min}\limits_{\widehat{v_{\tau}} \in \widehat{P_{\tau}}} \left \| v_{\tau} - \widehat{v_{\tau}}\right \|_{2}$ enforces that any 3D point $v_{\tau}$ in temporally colorized skeleton cloud $P_{\tau}$ has a matched 3D point $\widehat{v_{\tau}}$ in the repainted point cloud $\widehat{P_{\tau}}$. The term $\mathop{\min} \limits_{v_{\tau} \in P_{\tau}} \left \| \widehat{v_{\tau}} - v_{\tau} \right \|_{2}$ enforces the matching vice versa. The max operation enforces that the distance from $P_{\tau}$ to $\widehat{P_{\tau}}$ and vice versa need to be small concurrently. 

By using the Chamfer distance, the encoder $E_{\tau}(.)$ and decoder $D_{\tau}(.)$ are forced to recover temporal color features for points $v_{r}$ in the raw skeleton cloud $P_{r}$. Similarly, the encoder $E_{s}(.)$ and decoder $D_{s}(.)$ will learn spatial color features, and the encoder $E_{p}(.)$ and decoder $D_{p}(.)$ are pushed to distinguish the person index and learn interactive information. 

\noindent\textbf{Two-Stream Pipeline:}
As mentioned in Section~\ref{sec: C-F}, we introduce two coarse-grained colorization clouds ($P_{\tau c}$ and $P_{sc}$) which capture the coarse-grained temporal dependency and spatial relation information. 
As shown in Fig.~\ref{fig:align model temporal} and Fig.~\ref{fig:align model spatial}, together with the original temporal and spatial colorizations, we design a two-stream auto-encoder framework for learning both the fine-grained and coarse-grained spatial-temporal information. 
{Additionally, we do the feature alignment at the latent feature level to align the representations learned from varying spatial and temporal colorization granularities. This alignment guides the fine-grained encoders ($E_{\tau}$ and $E_{s}$) in capturing both fine and coarse spatial-temporal information.}

Using Fig.~\ref{fig:align model temporal} as an example, we utilize the Chamfer Distance to measure the repainting error both between the ground truth temporal colorization $P_{\tau}$ and the repainted $\widehat{P_{\tau}}$, and between the ground truth temporal coarse-grained colorization $P_{\tau c}$ and the repainted $\widehat{P_{\tau c}}$.
{For feature alignment, we apply the Mean Squared Error (MSE).
The latent features for two stream are defined as $F_{\tau} = E_{\tau} (P_{\tau})$ and $F_{\tau c} = E_{\tau c} (P_{\tau c})$, respectively. 
The feature alignment loss, denoted by $L_{fa}$, is determined using the equation:
\begin{equation}
    L_{fa} = \frac{1}{n} \sum_{i=1}^{n} (P_{\tau c}^{i} - P_{\tau}^{i}),
\end{equation}
}
where $n$ indicates the dimension of the latent features. 
Thus, the final self-supervised models comprise both these two coarse-fine colorization alignment frameworks and the single-stream person-level colorization models.

\begin{table}[t]
\begin{center}
\caption{Comparisons of different network configurations' results with unsupervised and supervised settings on NTU RGB+D dataset.(`\textit{TS}': Temporal Stream; `\textit{SS}': Spatial Stream; `\textit{PS}': Person Stream; `3s' means three-stream fusion)
}
\label{tab: abla_unsup_and_sup}
\begin{tabular}{|l|c|c|} 
  \hline 
  Dataset &  NTU-CS  &  NTU-CV  \\
  \hline
  \hline
  \multicolumn{3}{|c|}{ \textbf{Unsupervised Setting}} \\
  \hline
  \hline
  Baseline-U &    $61.8$     &    $68.4$             \\
  Motion Prediction-U & $65.7$    &  $75.0$            \\
  Masked Autoencoder-U & $65.9$    &  $75.3$         \\
  \hline
  `\textit{TS}' Colorization   &    $71.6$       &     $90.1$      \\
  `\textit{SS}' Colorization   &   $68.4$    &    $77.5$        \\
  `\textit{PS}' Colorization   &    $64.2$    &      $72.8$     \\
  \hline
  `\textit{TS} $+$ \textit{SS}' Colorization &  $74.6$  &   $82.6$      \\
  `\textit{TS} $+$ \textit{PS}' Colorization &  $73.3$   &     $81.4$   \\
  `\textit{SS} $+$ \textit{PS}' Colorization &   $69.6$   &    $78.6$    \\
  \hline
  3s-Colorization &  \textbf{75.2}    &  \textbf{83.1}     \\
  \hline
  \hline
  \multicolumn{3}{|c|}{ \textbf{Supervised Setting}} \\
  \hline
  \hline
  Baseline-S &    $76.5$     &   $83.4$             \\
  Motion Prediction-S &    $76.6$   &   $83.9$            \\
  Masked Autoencoder-S &    $77.4$     &    $84.3$             \\
  \hline
  `\textit{TS}' Colorization &   $84.2$     &  $93.1$           \\
  `\textit{SS}' Colorization    &   $82.3$    &   $91.5$           \\
  `\textit{PS}' Colorization   &   $81.1$     &     $90.3$     \\
  \hline
  `\textit{TS} $+$ \textit{SS}' Colorization &  $86.3$   &  $94.2$       \\
  `\textit{TS} $+$ \textit{PS}' Colorization &  $86.4$  &    $94.1$    \\
  `\textit{SS} $+$ \textit{PS}' Colorization &  $85.0$     &   $93.0$     \\
  \hline
    3s-Colorization & \textbf{88.0}    &   \textbf{94.9}    \\
  \hline
\end{tabular}
\end{center}
\end{table}

\begin{table*}[h]
\begin{center}
\caption{Comparisons of different network configurations' results with the semi-supervised setting on NTU RGB+D dataset. (`\textit{TS}': Temporal Stream; `\textit{SS}': Spatial Stream; `\textit{PS}': Person Stream; `3s' means three-stream fusion; and the number in parentheses denotes the number of labeled samples per class)
}
\label{tab:abla semi-supervised result NTU}
\resizebox{0.95\textwidth}{!}{
\begin{tabular}{l|c|c|c|c|c|c|c|c|c|c}
  \hline 
  \multirow{2}{*}{Method} & \multicolumn{2}{c|}{ \textbf{$1\%$}}&\multicolumn{2}{c|}{ \textbf{$5\%$}}& \multicolumn{2}{c|}{ \textbf{$10\%$}}& \multicolumn{2}{c|}{ \textbf{$20\%$}}& \multicolumn{2}{c}{ \textbf{$40\%$}}\\ 
  \cline{2-11} 
   & CS $(7)$ & CV $(7)$& CS $(33)$& CV $(31)$ & CS $(66)$ & CV $(62)$ & CS $(132)$ & CV $(124)$ & CS $(264)$ & CV $(248)$ \\ 
  \hline
  Baseline-Semi  &  27.1   & 28.1 & 46.0 & 50.6 & 55.1 & 60.7 & 60.9 & 69.1 & 64.2 & 73.7 \\
  Motion prediction-Semi   &  36.2   & 38.3 & 52.9 & 56.2 & 58.5  & 64.4 & 64.0 & 70.8  & 69.3 & 76.7 \\
  Masked autoencoder-Semi   &  35.5   & 38.6 & 52.4 & 55.6 & 59.4 & 64.3  & 63.9  & 71.7 & 69.5 & 76.9 \\
  
  \hline
   `\textit{TS}' Colorization          & 42.9   & 46.3 & 60.1 & 63.9 & 66.1 & 73.3 & 72.0 & 77.9 & 75.9 & 82.7 \\ 
  `\textit{SS}' Colorization                     & 40.2   & 43.1 & 54.6 & 60.0 & 60.1 & 68.1 & 64.2 & 73.1 & 69.1 & 77.6 \\
  `\textit{PS}' Colorization                     & 37.9   & 40.1 & 51.2 & 56.0 & 56.8 & 63.2 & 61.9 & 70.2 & 65.8 & 74.6 \\
  \hline
  `\textit{TS} $+$ \textit{SS}' Colorization  & 48.1   & 51.5 & 64.7 & 69.3 & 70.8 & 78.2 & 75.2 & 81.8 & 79.2 & 86.0 \\
  `\textit{TS} $+$ \textit{PS}' Colorization & 46.9   & 50.5 & 63.9 & 68.1 & 69.8 & 77.0 & 74.9 & 81.3 & 78.3 & 85.4 \\
  `\textit{SS} $+$ \textit{PS}' Colorization & 43.2   & 46.7 & 58.3 & 64.4 & 64.2 & 72.0 & 69.0 & 77.1 & 73.2 & 81.5 \\
  \hline
  3s-Colorization & \textbf{48.3}  & \textbf{52.5} & \textbf{65.7} & \textbf{70.3} & \textbf{71.7} & \textbf{78.9} & \textbf{76.4} & \textbf{82.7} & \textbf{79.8} & \textbf{86.8} \\
  \hline
\end{tabular}
}
\end{center}
\end{table*}


\begin{table}[t]
\caption{
Ablation studies on the Segment Size and Body Part Scale. The experiments conduct on NTU RGB+D Dataset under the unsupervised setting.
}
\label{tab: Segment Body}
\begin{center}
\resizebox{0.49\textwidth}{!}{
\begin{tabular}{|ccc|ccc|}
  \hline 
  Segment Size &  NTU-CS & NTU-CV  & Spatial scale & NTU-CS & NTU-CV \\
  
  \hline
  2  & 72.2  & 82.3  & 1 (10 parts) & \textbf{72.5} & \textbf{82.1}  \\
  4  & 72.8  & 82.2  & 2 (6 parts)  & 71.3 & 81.9 \\
  5  & \textbf{73.2}  & \textbf{82.6}  & & & \\
  8  & 73.1  & 82.5   & & & \\
  10 & 72.5  & 81.8 &  & &  \\
  20 & 72.5  & 82.1   & & &\\
  \hline
\end{tabular}
}
\end{center}
\end{table}

\subsection{Masked Skeleton Cloud Modeling}
\label{sec: mask}
Repainting $P_{r}$ to colorized skeleton clouds is a non-trivial task, requiring a balance between color repainting and self-supervised feature learning.  
Motivated by BERT~\cite{devlin2018bert} and MAE~\cite{he2022masked}, we study the masked skeleton cloud modeling strategy for skeleton representation learning based on the introduced auto-encoder framework.
Specifically, we devise a Masked Skeleton Cloud Repainting task that aims at reconstructing the geometric structure and color information of the masked points from partially visible points.
The proposed task is based on the introduced auto-encoder, where an encoder is utilized to map the visible skeleton points, together with the corresponding color information, into latent representations, and the decoder reconstructs the geometric information and the corresponding color information from the latent representations.   
By modeling the masked parts, the model attains the spatial-temporal understanding of the skeleton sequence.
Below, we provide more details of the masking strategies designed for the colorized skeleton cloud.

\noindent\textbf{Mask Sampling Strategy.} 
There are five masking strategies: random masking, frame-only masking, segment masking, joint-only masking, and body-part masking, as shown in Fig.~\ref{fig:masking}.
(1) \textit{random masking}, we randomly and uniformly select a subset of skeleton points. 
For the reason that the temporal correlation and joint interaction are critical for skeleton action recognition, we design the specific masking strategy based on the temporal and spatial dimension (Fig.~\ref{fig:masking} (b) and (f)).  
(2) \textit{frame-only masking}, we mask a subset of skeleton points, which belong to randomly selected frames (Fig.~\ref{fig:masking} (c)).  
(3) \textit{segment masking}, we randomly choose a location as the center of the sampled sequence and mask all points from the selected continuous frames (Fig.~\ref{fig:masking} (d)).  
(4) \textit{joint-only masking}, we mask a subset of skeleton points, which belong to randomly selected joints (Fig.~\ref{fig:masking} (g)).  
(5) \textit{body-part masking}, we mask a subset of skeleton points, which belong to randomly selected body parts (Fig.~\ref{fig:masking} (h)). 

\noindent {\textbf{Implementation of masking strategy.} 
For the masked points, we initial all the position and color values to zeros. Consequently, The unmasked point is represented by $[x, y, z, r, g, b]$,
while the masked point is initialed as $[0, 0, 0, 0, 0, 0]$.
Note that the masked points are randomly selected for every sample during each epoch.}

\subsection{Training objectives}
\label{training}
\noindent\textbf{Self-Supervised Repainting.} 
In this stage, we aim to repaint from three perspectives. For temporal and spatial colorization, the training objective contains two reconstruction losses (Chamfer distance) on fine-grained and coarse-grained colorization levels and alignment loss (mean squared error, MSE) between latent features, as shown in Fig. \ref{fig:align model temporal} and Fig. \ref{fig:align model spatial}.
For person-level colorization, since there is no coarse-grained colorization, the training objective is only one Chamfer distance, as shown in Fig.~\ref{fig:framework}.

\noindent\textbf{Skeleton Action Recognition.}
As described in Section~\ref{Pipeline}, we introduce the alignment loss to focus the latent feature of each auto-encoder to contain both coarse and fine spatial-temporal information. 
To maintain the computation cost the same as that of \cite{Yang_2021_ICCV}, we only use the fine-grained encoders (\textit{i.e.}, $E_{\tau}(.)$, $E_{s}(.)$) in this stage. 
Together with the $E_{p}(.)$, we obtain three encoders that capture meaningful temporal, spatial, and interaction features, respectively. 
With the feature representations from the three encoders, we include a simple linear classifier $f(.)$ on top of the encoder to perform action recognition as in~\cite{kundu2019unsupervised, lin2020ms2l, nie2020view, su2020predict, zheng2018unsupervised}. We adopt different settings to train the classifier, including unsupervised, semi-supervised, and supervised settings. In the unsupervised setting, the encoder is only trained by the skeleton cloud repainting method, and then we train the linear classifier with the encoder fixed by following previous unsupervised skeleton representation learning works \cite{kundu2019unsupervised, lin2020ms2l, nie2020view, su2020predict, zheng2018unsupervised}. In the semi-supervised and supervised settings, the encoder is first trained with self-supervised representation learning and then fine-tuned with the linear classifier as in \cite{lin2020ms2l,si2020adversarial}. We use the standard cross-entropy loss as the classification loss $L_{cls}$.

\section{Experiments}
\label{sec: experiment}
{We conducted extensive experiments over five publicly accessible datasets, including NTU RGB+D \cite{Shahroudy_2016_CVPR}, NTU RGB+D 120~\cite{ntu120}, PKU-MMD \cite{pkummd}, Northwestern-UCLA \cite{wang2014cross}, and UWA3D \cite{rahmani2014hopc}.} The experiments aim to evaluate whether our skeleton cloud colorization scheme can learn effective self-supervised feature representations for the task of skeleton action recognition. We therefore evaluate different experimental settings, including unsupervised, semi-supervised, and supervised, as well as transfer learning.

\begin{table}[t]
\begin{center}
\caption{Linear evaluation results compared with Skeleton Colorization~\cite{Yang_2021_ICCV} on NTU-60 dataset. `$\Delta$' represents the gain compared to~\cite{Yang_2021_ICCV} with the same stream data. (C-F stands for coarse-fine; `\textit{TS}': Temporal Stream; `\textit{SS}': Spatial Stream; `\textit{PS}': Person Stream; `3s' means three-stream fusion)
}
\label{tab: Cosine-to-Fine}
\begin{tabular}{|l|cc|cc|}
  \hline 
  \multirow{3}{*}{Method} &  \multicolumn{4}{c|}{\textbf{NTU RGB+D}}  \\
  \cline{2-5} 
  & \multicolumn{2}{c|}{C-Subject} &  \multicolumn{2}{c|}{C-View}   \\
  & Acc. & $\Delta$  & Acc. & $\Delta$ \\
  \hline
  `\textit{TS}' Colorization \cite{Yang_2021_ICCV}  & 71.6 & & 79.9 &   \\
  `\textit{TS}' Masked Colorization    & 72.1 & $\uparrow$ 0.5 & 81.1 &  $\uparrow$ 1.2    \\
  `\textit{TS}' C-F Masked Colorization  (Ours)    & 73.2 & $\uparrow$ 1.6 & 82.6 & $\uparrow$ 2.7 \\
  \hline
  `\textit{SS}' Colorization \cite{Yang_2021_ICCV}  & 68.4 & & 77.5 &  \\
  `\textit{SS}' Masked Colorization   & 69.6 & $\uparrow$ 1.2 & 80.1&   $\uparrow$ 2.6     \\
  `\textit{SS}' C-F Masked Colorization  (Ours)     & 72.5 &$\uparrow$ 4.1 & 82.1 & $\uparrow$ 4.6    \\
  \hline
  `\textit{PS}' Colorization \cite{Yang_2021_ICCV}  & 64.2 & & 72.8 &  \\
  `\textit{PS}' Masked Colorization  & 67.9   & $\uparrow$ 3.7 & 77.1 &  $\uparrow$ 4.3   \\
  \hline
  3s-Colorization  \cite{Yang_2021_ICCV} & 75.2 & & 83.1 &   \\
  3s-Masked Colorization  &77.2 & $\uparrow$ 2.0  & 85.8 & $\uparrow$ 2.7  \\
  \textbf{3s-C-F Masked Colorization (Ours)}  & \textbf{79.1} &  $\uparrow$ 3.9  & \textbf{87.2} & $\uparrow$ 4.1 \\
  \hline
  
\end{tabular}
\end{center}
\end{table}

\begin{table*}[t]
\begin{center}
\caption{Ablation study on temporal masking strategy, including the temporal random masking (Fig.~\ref{fig:masking} (b)), frame-only masking (Fig.~\ref{fig:masking} (c)), and segment masking (Fig.~\ref{fig:masking} (d)). The experiments conduct on NTU RGB+D under the unsupervised setting.
}
\label{tab:temporal masking}
\setlength\tabcolsep{2pt}
\begin{tabular}{|c c c|c c c|c c c|}
  \hline 
\multicolumn{3}{|c|}{ \textbf{(b) Random Masking}} & \multicolumn{3}{c|}{ \textbf{(c) Frame-only Masking}} & \multicolumn{3}{c|}{ \textbf{(d) Segment Masking}} \\
  \hline
    \hline
    Mask Ratio & NTU-CS  & NTU-CV & Mask Frame Number & NTU-CS  & NTU-CV & Mask Segment Length & NTU-CS  & NTU-CV \\
  \hline

    0.25 & 73.2 & 82.6 &  5 & 71.7 & 82.4 &  5 & 72.8 & 82.6 \\
    0.50 & 72.8 & 83.1 & 10 & 73.4 & 82.7 & 10 & 73.5 & 83.1 \\
    0.75 & 72.8 & 82.8 & 15 & 72.8 & 83.1 & 15 & \textbf{74.0} & \textbf{83.5} \\
         &        &   &  20 & 73.4 & 82.8   & 20 & 73.4 & 83.3  \\
         &        &   &  30 & 73.5 & 82.3   & 30 & 73.8 & 83.3 \\
  \hline
\end{tabular}
\end{center}
\end{table*}

\begin{table*}[t]
\begin{center}
\caption{Ablation study on spatial masking strategy, including the spatial random masking (Fig.~\ref{fig:masking} (f)), joint-only masking (Fig.~\ref{fig:masking} (g)), and body-part masking (Fig.~\ref{fig:masking} (h)). The experiments conduct on NTU RGB+D under the unsupervised setting.
}
\label{tab:spatial masking}
\setlength\tabcolsep{2pt}
\begin{tabular}{|c c c|c c c|c c c|}
  \hline 
\multicolumn{3}{|c|}{ \textbf{(f) Random Masking}} & \multicolumn{3}{c|}{ \textbf{(g) Joint-only Masking}} & \multicolumn{3}{c|}{ \textbf{(h) Body-part Masking}} \\
  \hline
  \hline 
    Mask Ratio & NTU-CS  & NTU-CV  & Mask Joint Number & NTU-CS  & NTU-CV & Mask Part Number & NTU-CS  & NTU-CV \\
  \hline
   0.25 & 72.5 & 82.1   & 5 & 72.5 & 82.3  & 2 & 72.2 & 82.2\\
   0.50 & 72.4  & 81.9    & 10 & \textbf{72.8} & \textbf{82.6}   & 4 & 72.2 & 82.3 \\
   0.75 & 71.7 & 81.4 &  15 & 72.3 & 82.3   & 6 & 72.0 & 82.4 \\
      &      &   & 20 & 72.5 & 82.2   & 8 & 72.0 & 81.7 \\
  \hline
\end{tabular}
\end{center}
\end{table*}

\subsection{Datasets}
{\bf NTU RGB+D \cite{Shahroudy_2016_CVPR}.} NTU RGB+D consists of 56880 skeleton action sequences which is the most widely used dataset for skeleton-based action recognition research. In this dataset, action samples are performed by 40 volunteers and categorized into 60 classes. Each sample contains an action and is guaranteed to have at most two subjects as captured by three Microsoft Kinect v2 cameras from different views. The authors of this dataset recommend two benchmarks: (1) cross-subject (CS, C-Subject) benchmark, where training data comes from 20 subjects and testing data comes from the other 20 objects; (2) cross-view (CV, C-View) benchmark, where training data comes from camera views 2 and 3, and testing data comes from camera view 1.

\noindent{\bf NTU RGB+D 120 \cite{ntu120}.} 
NTU RGB+D 120 dataset is currently the largest dataset with 3D joint annotation for human action recognition. The dataset contains 114480 action samples in 120 action classes. Samples are captured by 106 volunteers with three camera views. This dataset contains 32 setups, and each setup denotes a specific location and background. 
The author of this dataset recommends two benchmarks: (1) cross-subject (C-Subject) benchmark: training data comes from 53 subjects, and the testing data comes from the other 53 objects. (2) cross-setup (C-Setup) benchmark: training data comes from samples with even setup IDs, and testing data comes from samples with odd setup IDs.

\noindent{\bf PKU-MMD \cite{pkummd}.} 
PKU-MMD is a new large-scale benchmark for continuous multi-modality 3D human action understanding and covers a wide range of complex human activities with well-annotated information.
It contains almost 20, 000 action instances and 5.4 million frames in 51 action categories.
Each sample consists of 25 body joints. 
PKU-MMD consists of two subsets, \textit{i.e.}, parts I and II. Part I is an easier version for skeleton action recognition, while part II is more challenging, with more skeleton noise caused by the large view variation. We conduct experiments under the cross-subject protocol on the two subsets respectively.

\noindent{\bf Northwestern-UCLA (NW-UCLA) \cite{wang2014cross}.} 
This dataset is captured by three Kinect v1 cameras, and it contains 1494 samples performed by 10 subjects. It contains 10 action classes, and each body has 20 skeleton joints. Following the evaluation protocol in \cite{wang2014cross}, the training set consists of samples from the first two cameras ($V1$, $V2$) and the rest of the samples from the third camera ($V3$) form the testing set.

\noindent{\bf Multiview Activity II (UWA3D) \cite{rahmani2014hopc}.} 
UWA3D dataset contains 30 human actions performed 4 times by 10 subjects. 15 joints are recorded, and each action is observed from four views: front (V1), left side (V2), right sides (V3), and top view (V4). The total number of action sequences is 1075. The dataset is challenging due to many views and self-occlusions.

\noindent{{\bf 
Toyota Smarthome (Smarthome)~\cite{Das_2019_ICCV}.}
Toyota Smarthome is a real-world dataset for daily living action classification and contains 16,115 videos of 31 classes. 
We conduct the transfer learning experiments following the cross-subject (CS) and cross-view1 (CV1) evaluation protocols.}

Note that NW-UCLA, UWA3D, and Smarthome datasets only contain single-person actions, so we do not conduct person-level colorization experiments on these three datasets.


\subsection{Implementation Details}
\label{sec: implentation}
For NTU RGB+D, NTU RGB+D 120, and PKU-MMD, we adopt the pre-processing in \cite{Shi_2019_CVPR_twostream} and uniformly sampled $T = 40$ frames from each skeleton sequence. The sampled skeleton sequences are constructed to a 2000-point skeleton cloud. For NW-UCLA, we adopt the pre-processing in \cite{su2020predict} and uniformly sampled $T = 50$ frames from the skeleton sequences, and the skeleton cloud has 1000 points. For UWA3D, we also adopt the pre-processing in \cite{su2020predict} and uniformly sampled $T = 70$ frames from the skeleton sequences, and the skeleton cloud has 1050 points.

In the self-supervised feature learning phase, we use the Adam optimizer and set the initial learning rate at 1e-5, and reduce it to 1e-7 with cosine annealing. The dimension of the encoder output is 1024, and the batch size is 24 for all datasets. The training lasts for 150 epochs. 
{In the classifier training phase, the unmasked colored skeleton cloud is utilized as the input. We use the SGD optimizer with Nesterov momentum (0.9). We set the initial learning rate at 0.001 and reduce it to 1e-5 with cosine annealing. The batch size is 32 for all datasets, and the training lasts for 200 epochs. We implement our method using PyTorch, and all experiments were conducted on NVIDIA RTX 3090 GPUs with CUDA 11.3.}

\begin{table*}[t]
\begin{center}
\caption{Comparisons to state-of-the-art self-supervised skeleton action recognition method on NTU RGB+D, NTU RGB+D 120, PKU-MMD, UWA3D, and NW-UCLA datasets. (C-F stands for coarse-fine; `2s' means three-stream fusion; `3s' means three-stream fusion) 
}
\label{tab:unsupervised result NTU}
\resizebox{0.99\textwidth}{!}{
\begin{tabular}{|l|c|c|c|c|c|c|c|c|c|c|c|}
  \hline 
  \multirow{2}{*}{Method} & \multirow{2}{*}{Backbone} & \multirow{2}{*}{Stream} & \multicolumn{2}{c|}{\textbf{NTU RGB+D}} & \multicolumn{2}{c|}{\textbf{NTU RGB+D 120}} & \multicolumn{2}{c|}{\textbf{PKU-MMD}}  & \multicolumn{2}{c|}{\textbf{UWA3D}} & \multirow{2}{*}{\textbf{NW-UCLA}} \\ 
  \cline{4-11} 
  & & & C-Subject  & C-View & C-Subject  & C-Setup & I  & II & V3 (\%) & V4 (\%) & \\
  \hline
  LongT GAN \cite{zheng2018unsupervised} (AAA1 2018)& GRU & 1& 39.1 & 52.1 & 35.6 & 39.7 & 68.7 & 26.5 & 53.4 & 59.9 & 74.3 \\
  P\&C FS-AEC \cite{su2020predict} (CVPR 2020)& GRU &1 & 50.6 & 76.3 & -- & -- & -- & -- & 59.5 & 63.1 & 83.8\\ 
  P\&C FW-AEC \cite{su2020predict} (CVPR 2020)& GRU &1 & 50.7 & 76.1  & 41.1 & 44.1 & 59.9 & 25.5 & 59.9 & 63.1 & 84.9\\ 
  M$S^{2}$L \cite{lin2020ms2l} (ACMMM 2020)& GRU & 1 & 52.6  & -- & -- & -- & 64.9    & 27.6 &-- & -- & 76.8\\ 
  PCRP \cite{xu2021prototypical} (TMM 2021)& GRU & 1 & 53.9  & 63.5 & 41.7  & 45.1 & -- & -- & -- & -- & 87.0\\
  AS-CAL \cite{rao2021augmented} (Information Sciences 2021)& LSTM & 1 & 58.5 & 64.8 & -- & -- & -- & -- & -- & -- & --\\
  GLTA-GCN \cite{9859752} (ICME 2022)& GCN & 2 & 61.2 & 81.2 & 49.1 & 51.1 & -- & -- & 61.5 & 68.2 & -- \\
  MCAE-MP \cite{xu2021unsupervised} (NeurIPS 2021)& MCAE & 1 & 65.6  & 82.4 & 52.8  & 54.7 & -- & -- & -- & -- & 84.9\\
  Taxonomy-SSL \cite{tanfous2022and} (WACV 2022) & GRU & 1 & 67.0 & 76.3 & 59.1 & 61.5 & --& -- & -- & -- & 86.1\\  
  CRRL~\cite{9901454} (TIP 2022)& GRU &  1 & 67.6 & 73.8  & 57.0 &  56.2 & -- & -- & -- & -- & 83.8 \\ 
  ST-CL($\times 4$) \cite{gao2021efficient} (TMM 2021)& GCN & 4 & 68.1   & 69.4 & 54.2   & 55.6 & -- & --  & 46.0  & 44.0 & 81.2\\
  
  EnGAN-PoseRNN \cite{kundu2019unsupervised} (WACV 2019)& RNN & 3 &68.6 &  77.8 & -- & -- & -- & -- & -- & -- & --\\
  H-Transformer \cite{cheng2021hierarchical} (ICME 2021)& Transformer & 1 & 69.3 & 72.8 & -- & -- & -- & -- & -- & -- & 83.9\\
  CP-STN \cite{zhan2021spatial} (ACML 2021)& GCN & 1 & 69.4 & 76.6  & 55.7 & 54.7 & -- & -- & -- & -- & --\\ 
  
  SeBiReNet \cite{nie2020view} (ECCV 2020)& GRU& 2   & -- & 79.7 & -- &  69.3 & -- & -- & 53.9 &  61.6 & 80.3\\
  SKT \cite{zhang2022skeletal} (ICME 2022)& GCN & 1 & 72.6 & 77.1 & 62.6 & 64.3 & --& -- & -- & -- & --\\ 

  2s-Colorization \cite{Yang_2021_ICCV} (ICCV 2021) & DGCNN & 2         &  -- &  -- & -- & -- & --& -- & 70.0 & 70.6 & 91.1\\
  3s-Colorization \cite{Yang_2021_ICCV} (ICCV 2021) & DGCNN &  3        &  75.2 &  83.1 & 64.3  & 67.5  & 87.2 & 47.1 & --& -- &--\\
  GL-Transformer \cite{GL-Transformer} (ECCV 2022) & Transformer & 1& 76.3 & 83.8 & 66.0 &  68.7 & -- & -- & -- & -- & 90.4\\
  Skeleton-Contrastive  \cite{thoker2021skeleton} (ACMMM 2021) & GRU+CNN & 2& 76.3 & 85.2 & 67.1 & 67.9 & 80.9 & 36.0 & -- & -- & --\\
  3s-CrosSCLR \cite{Li_2021_CVPR} (CVPR 2021) & GCN & 3 & 77.8 & 83.4 & 67.9 & 66.7 & 84.9 & 21.1 & -- & -- &--\\
  3s-HiCLR \cite{zhang2022hierarchical} (AAAI 2023) & GCN & 3 & 78.8  & 83.1 & 67.3  & 69.9 & --& -- & -- & -- & --\\ 
  3s-AimCLR \cite{guo2022contrastive} (AAAI 2022) & GCN & 3 & 78.9 & 83.8 & 68.2 & 68.8 & -- & -- & -- & -- & -- \\

  \hline
  \textbf{2s-C-F Masked Colorization (Ours)}  & DGCNN &     2 &   -- & -- &  -- & -- & --  &   -- & \textbf{71.7} & \textbf{73.8} & \textbf{92.0}\\ 
  \textbf{3s-C-F Masked Colorization (Ours)}  & DGCNN &    3  &   \textbf{79.1} & \textbf{87.2} &  \textbf{69.2} & \textbf{70.8} & \textbf{89.2}  &   \textbf{49.8} & -- & -- & --\\ 
  \hline
\end{tabular}
}
\end{center}
\end{table*}

\begin{table*}[t]
\caption{Comparisons of action recognition results with semi-supervised learning approaches on NTU RGB+D dataset. (C-F stands for coarse-fine; `3s' means three-stream fusion; * means the reproduced results with our labeled/unlabeled splitting; and the number in parentheses denotes the number of labeled samples per class )
}
\begin{center}
\resizebox{0.99\textwidth}{!}{
\begin{tabular}{|l|c|c|c|c|c|c|c|c|c|c|c|c|}
  \hline 
  \multirow{2}{*}{Method} & \multirow{2}{*}{Backbone} & \multirow{2}{*}{Streams} & \multicolumn{2}{c|}{ \textbf{$1\%$}}&\multicolumn{2}{c|}{ \textbf{$5\%$}}& \multicolumn{2}{c|}{ \textbf{$10\%$}}& \multicolumn{2}{c|}{ \textbf{$20\%$}}& \multicolumn{2}{c|}{ \textbf{$40\%$}}\\ 
  \cline{4-13} 
  &  & & CS $(7)$ & CV $(7)$& CS $(33)$& CV $(31)$ & CS $(66)$ & CV $(62)$ & CS $(132)$ & CV $(124)$ & CS $(264)$ & CV $(248)$ \\ 
  \hline
  $ S^{4}L$(Inpainting) \cite{zhai2019s4l} (ICCV 2019)& GRU & 1 & --   & -- & 48.4 & 55.1 & 58.1 & 63.6 & 63.1 & 71.1 & 68.2 & 76.9 \\
  Pseudolabels  \cite{lee2013pseudo} (ICML 2013) & GRU & 1      & --   & -- & 50.9 & 56.3 & 57.2 & 63.1 & 62.4 & 70.4 & 68.0 & 76.8 \\
  VAT    \cite{miyato2018virtual} (TPAMI 2018)  &  GRU & 1           & --   & -- & 51.3 & 57.9 & 60.3 & 66.3 & 65.6 & 72.6 & 70.4 & 78.6 \\
  VAT + EntMin \cite{grandvalet2005semi} (NeurIPS 2005) &  GRU & 1       & --   & -- & 51.7 & 58.3 & 61.4 & 67.5 & 65.9 & 73.3 & 70.8 & 78.9 \\
  
  ASSL \cite{si2020adversarial} (ECCV 2020) & GRU & 1 & --   & -- & 57.3 & 63.6 & 64.3 & 69.8 & 68.0 & 74.7 & 72.3 & 80.0 \\
  M$S^{2}$L \cite{lin2020ms2l} (ACMMM 2020) & GRU  & 1 & 33.1 & -- & --   & --   & 65.2 & --   & --   & --   & --   & --\\
  LongT GAN \cite{zheng2018unsupervised} (AAAI 2018)& GRU  &1  & 35.2 & -- & --   & --   & 62.0 & --   & --   & --   & --   & --\\
  Skeleton-Contrastive \cite{thoker2021skeleton} (ACMMM 2021) & GRU+CNN  & 4 & 35.7 & 38.1 & 59.6   & 65.7   & 65.9 & 72.5   & 70.8   & 78.2   & --   & --\\
  AL+K \cite{li2020sparse} (arxiv 2020)  & GRU  &  1 & 41.8 &  -- & 57.8   & --   & 62.9 & --  & --   & --   & --   & --\\
  SKT \cite{zhang2022skeletal} (ICME 2022)& GCN & 1 & 43.2 &  44.9 & --   & --   & 67.6 & 71.3  & --   & --   & --   & --\\
  MAC \cite{9954217} (TPAMI 2022)& GCN  & 2 & -- & -- & 63.3   & 70.4   & 74.2 & 78.5  & 78.4   & 84.6   & 81.1   & 89.6\\
  GL-Transformer \cite{GL-Transformer} (ECCV 2022)& Transformer &  1 & -- &  -- & 64.5   & 68.5   & 68.6 &  74.9  & --   & --   & --   & --\\
  3s-Colorization \cite{Yang_2021_ICCV} (ICCV 2021)& DGCNN & 3 & 48.3 & 52.5 & 65.7 & 70.3 & 71.7 & 78.9 & 76.4 & 82.7 & 79.8 & 86.8 \\
  \hline
  3s-CrosSCLR \cite{Li_2021_CVPR}* (CVPR 2021)& GCN & 3 & 49.9$\pm$1.30   & 52.0$\pm$0.48 & 66.5$\pm$0.50  & 69.5$\pm$1.55   &  73.3$\pm$0.43 &  77.4$\pm$0.30 &  77.0$\pm$0.33  & 82.9$\pm$0.14   & 81.0$\pm$0.30   &  87.3$\pm$0.20 \\
  3s-AimCLR \cite{guo2022contrastive}* (AAAI 2022)& GCN & 3 & 45.5$\pm$1.72  &  46.9$\pm$2.08 & \textbf{68.1$\pm$0.48} &  71.8$\pm$0.80  &  76.1$\pm$0.43 & 80.7$\pm$0.71  & 77.8$\pm$0.38   &  84.0$\pm$0.29  &  81.5$\pm$0.51  & 88.6$\pm$0.28 \\
  \hline
  \textbf{3s-C-F Masked Colorization (Ours)}& DGCNN & 3 & \textbf{52.3$\pm$0.57}  & \textbf{53.1$\pm$0.97} & \textbf{68.1$\pm$0.19} & \textbf{74.2$\pm$1.05} & \textbf{76.5$\pm$0.27} & \textbf{81.3$\pm$0.30} & \textbf{78.7$\pm$0.27} & \textbf{85.7$\pm$0.46} & \textbf{82.5$\pm$0.28} & \textbf{89.7$\pm$0.10} \\
  \hline
\end{tabular}
}
\end{center}

\label{tab:semi-supervised result NTU}
\end{table*}

\begin{table*}[t]
\caption{Comparisons of action recognition results with semi-supervised learning approaches on NW-UCLA dataset. (C-F stands for coarse-fine; `2s' means two-stream fusion; $v./c.$ denotes the number of labeled videos per class)
}
\begin{center}
\resizebox{0.99\textwidth}{!}{
\begin{tabular}{|l|c|c|c|c|c|c|c|c|}
  \hline 
  Method & Backbone & Stream & $1\% \; (1 \;v_{.}/c_{.})$ &  $5\% \; (5 \;v_{.}/c_{.})$ &  $10\% \; (10 \;v_{.}/c_{.})$ &  $15\% \; (15 \;v_{.}/c_{.})$ &  $30\% \; (30 \;v_{.}/c_{.})$ &  $40\% \; (40 \;v_{.}/c_{.})$  \\
  \hline
  $ S^{4}L$(Inpainting) \cite{zhai2019s4l} (ICCV 2019)  & GRU & 1 &  --    & 35.3 &  --    & 46.6  & 54.5  & 60.6\\
  Pseudolabels \cite{lee2013pseudo} (ICML 2013)    & GRU & 1       &  --    & 35.6 &  --    & 48.9  & 60,6  & 65.7 \\
  VAT      \cite{miyato2018virtual} (TPAMI 2018)   & GRU & 1             &  --    & 44.8 &  --    & 63.8  & 73.7  & 73.9\\
  VAT + EntMin \cite{grandvalet2005semi} (NeurIPS 2005)   & GRU & 1        &  --    & 46.8 &  --    & 66.2  & 75.4  & 75.6\\
  ASSL  \cite{si2020adversarial} (ECCV 2020)   & GRU & 1  &  --    & 52.6 &  --    & 74.8 & 78.0 & 78.4\\
  LongT GAN \cite{zheng2018unsupervised} (AAAI 2018)  & GRU & 1 &  18.3  & --   & 59.9   & --   & --   & --  \\
  M$S^{2}$L \cite{lin2020ms2l} (ACMMM 2020)  & GRU & 1&  21.3  & --   & 60.5   & --   & --   & --  \\
  2s-Colorization \cite{Yang_2021_ICCV} (ICCV 2021) & DGCNN& 2 & 41.9  &    57.2    &    75.0    &    76.0   &  83.0   &   84.9 \\
  GL-Transformer \cite{GL-Transformer} (ECCV 2022)  & Transformer & 1  & --  &   58.5    &   74.3   &    --  &  --   &   -- \\
  AL+K \cite{li2020sparse} (arxiv 2020)  & GRU & 1 & --  &   \textbf{63.9}    &    --   &    76.8  &  80.3   &   85.0 \\
  MAC \cite{9954217} (TPAMI 2022) & GCN & 2 & --  &   63.0    &    --   &    \textbf{78.8}  &  79.9   &   81.6 \\

  \hline
  
  \textbf{2s-C-F Masked Colorization (Ours)}  & DGCNN & 2&  \textbf{42.8}  &    61.6    &    \textbf{76.7}    &    \textbf{78.8}    &  \textbf{84.8}    &   \textbf{86.6} \\
  \hline
\end{tabular}
}
\end{center}

\label{tab:semi-supervised result NWUCLA}
\end{table*}

\subsection{Ablation Study}
We conduct ablation studies on different datasets to verify the effectiveness of different components of our method.

\subsubsection{Effectiveness of Our Skeleton Colorization:}
We verify the effectiveness of our skeleton cloud colorization on all three learning settings, including unsupervised learning, semi-supervised learning, and fully-supervised learning.

We compare our method with three baselines: 1) \textit{Baseline-U}: it only trains the linear classifier and freezes the encoder, which is randomly initialized; 2) \textit{Baseline-Semi}: the encoder is initialized with random weight instead of pre-training by our self-supervised representation learning; 3) \textit{Baseline-S}: the same as \textit{Baseline-Semi}. We train the encoder and linear classifier jointly with action labels. The input for these three baselines is the raw skeleton cloud without color label information. 

{Additionally, rather than merely comparing with models that rely on a randomly initialized encoder, we incorporate two commonly used video self-supervised methods to pre-train the encoder, resulting in two stronger baselines:
1) \textit{Motion Prediction}: During self-supervised pre-training, we mask the points from the final 10 frames, allowing the network to predict the motion information of these frames.  
2) \textit{Masked Autoencoder}: we randomly mask 25\% points during the self-supervised pre-training and push the network to reconstruct the masked information. 
It's worth noting that both of these two strong baselines are conducted without color information.}

We also study three skeleton cloud colorization configurations: 1) \textit{`T-Stream' (TS)} that uses temporally colorized skeleton cloud as self-supervision; 2) \textit{`S-Stream' (SS)} that uses spatially colorized skeleton cloud as self-supervision; 3) \textit{`P-Stream' (PS)} that uses the person-level colorized cloud as self-supervision.
`2s' means the combination of temporal and spatial steams, and `3s' stands for the combination of temporal, spatial, and person streams.

Table \ref{tab: abla_unsup_and_sup}  and Table \ref{tab:abla semi-supervised result NTU} show experimental results. It can be seen that all three colorization strategies (\textit{i.e.,} temporal-level, spatial-level, and person-level) achieve significant performance improvement as compared with the baseline, demonstrating the effectiveness of our proposed colorization technique. While the person-level colorization stream does not perform as well as the other two streams on the NTU RGB+D, it improves the overall performance while collaborating with the other two.
Moreover, when compared with the two stronger baselines (i.e., motion prediction and masked autoencoder), our proposed color repainting strategy still clearly outperforms them by a larger margin.


\subsubsection{Selection of Segment Size and Body Part Scale}
As introduced in Section~\ref{sec: C-F}, the coarse-grained skeleton colorization is based on the segment level and body part level. 

\noindent\textbf{Segment Size.} Table~\ref{tab: Segment Body} shows the classification results of our method with different segment sizes and body part scales on the NTU RGB+D dataset under the unsupervised setting.  
For the segment size, we can see that assigning each 5 consecutive frames on temporal color is the most helpful for temporal feature learning.

\noindent\textbf{Body Part Scale.} 
The human skeleton can be divided into ten body parts \cite{li2020dynamic} (scale 1, \textit{i.e.,} Neck, Trunk, Right arm, Right hand, Left arm, Left hand, Right leg, Right foot, Left leg, Left foot) or six body parts \cite{du2015hierarchical} (scale 2, \textit{i.e.,} Torso, Right upper limb, Left upper limb, Right lower limb, and Left lower limb) based on the human physical structure. As shown in Table~\ref{tab: Segment Body}, the performance of coarse-grained spatial colorization with scale 1 (10 parts) is better. 

Therefore, we use the 5-frame coarse-grained temporal colorization and 10-body-part (scale 1) coarse-grained spatial colorization in the main experiments.

\begin{table}[t]
\caption{Comparisons of action recognition results with semi-supervised learning approaches on NTU RGB+D 120 dataset C-Subject protocol. (C-F stands for coarse-fine; `3s' means three-stream fusion) 
}
\begin{center}
\scriptsize
\resizebox{0.49\textwidth}{!}{
\begin{tabular}{|l|c|c|c|c|c|c|}
  \hline 
  Method &  $1\% \;$&  $5\% \;$ &  $10\% \; $ &  $20\% \; $ &  $40\% \; $ &  $50\% \;$  \\
  \hline
  AS-CAL\cite{rao2021augmented} (Information Sciences 2021) &&  --  &    42.3   &    --    &    -- & 52.6   \\
  
  CP-STN \cite{zhan2021spatial} (ACML 2021)  & \textbf{48.8} &  51.5  &   56.1    &    60.6    &    66.5 & 67.8 \\

  \hline  
  \textbf{3s-C-F Masked Colorization (Ours)} & 46.5 &  \textbf{55.0}  &    \textbf{62.2}    &    \textbf{67.3}    &    \textbf{72.3} &  \textbf{75.4}   \\
  \hline
  
  \hline
  
\end{tabular}
}
\end{center}

\label{tab:semi-supervised result NTU120}
\end{table}

\begin{table}[t]
\caption{Comparisons of action recognition results with semi-supervised learning approaches on UWA3D dataset. (C-F stands for coarse-fine; `2s' means two-stream fusion) 
}
\begin{center}
\scriptsize
\resizebox{0.49\textwidth}{!}{
\begin{tabular}{|l|c|c|c|c|}
  \hline 
  Method &  $5\% \;$ &  $10\% \; $ &  $20\% \; $ &  $50\% \; $  \\
  \hline
  AL \cite{li2020sparse} (arxiv 2020)   & 26.9  &   37.1    &    41.2   &    55.8   \\
  AL+K \cite{li2020sparse} (arxiv 2020)  & 28.3  &   36.0    &    51.8   &    59.5   \\
  2s-Colorization \cite{Yang_2021_ICCV} (ICCV 2021) & 32.8  &   40.9    &    52.2    &    61.5  \\
  \hline
  \textbf{2s-C-F Masked Colorization (Ours)}  &  \textbf{36.5}  &    \textbf{44.2}    &    \textbf{54.5}    &    \textbf{63.8}     \\
  \hline
\end{tabular}
}
\end{center}

\label{tab:semi-supervised result UWA3D}
\end{table}

\subsubsection{Effectiveness of Masking Strategy and Coarse-Fine Alignment framework}
We conduct experiments on NTU RGB+D to verify the effectiveness of our Coarse-Fine Alignment framework and masking strategy under the unsupervised setting. All `masked' experiments in this ablation study are conducted with the random masking 25\%.

As shown in Table~\ref{tab: Cosine-to-Fine}, the masking task can improve the performance on all three colorization streams, and the proposed Coarse-Fine Alignment framework is able to cause further improvements, which demonstrates the benefit of the proposed masking strategy and coarse-fine Alignment framework.

\begin{table}[t]
\begin{center}
\footnotesize
\caption{Comparisons to state-of-the-art semi-supervised skeleton action recognition method on PKU-MMD dataset. (C-F stands for coarse-fine; `3s' means three-stream fusion)
}
\label{tab:semi-supervised result PKU}
\resizebox{0.49\textwidth}{!}{
\begin{tabular}{|l|c|c|c|c|}
  \hline 
  \multirow{2}{*}{Method} &  \multicolumn{2}{c|}{ \textbf{PKU-MMD part I}} &  \multicolumn{2}{c|}{ \textbf{PKU-MMD part II}}\\
  \cline{2-5} & Semi1 & Semi10  &  Semi1 &  Semi10 \\
  \hline
  LongT GAN \cite{zheng2018unsupervised} (AAAI 2018) & 35.8 &  69.5 & 12.4 & 25.7  \\
  M$S^{2}$L \cite{lin2020ms2l} (ACMMM 2021) & 36.4 &  70.3 & 13.0 & 26.1  \\
  Skeleton-Contrastive \cite{thoker2021skeleton} (ACMMM 2021) & 37.7 &  72.1 & -- & --  \\
  3s-CrosSCLR \cite{Li_2021_CVPR} (CVPR 2021) & 49.7 &  82.9 & 10.2 & 28.6  \\
  3s-AimCLR \cite{guo2022contrastive} (AAAI 2022) & 57.5 &  86.1 &  15.1 & 33.4  \\
  \hline
  \textbf{3s-C-F Masked Colorization (Ours)} & \textbf{57.8} & \textbf{86.5} & 
  \textbf{15.5} & \textbf{34.2}  \\
  \hline
 
\end{tabular}
}
\end{center}
\end{table}

\subsubsection{Effectiveness of Different Masking Strategies and Masking Ratios}
To find a proper masking strategy for our method, we conduct experiments on different masking types with different masking ratios on the temporal colorization and spatial colorization streams, respectively. This ablation study is conducted on the NTU RGB+D dataset under the unsupervised setting. The experimental results are presented in Table~\ref{tab:temporal masking} and Table~\ref{tab:spatial masking}.

It can be seen that the random masking does not work well in both scenarios. We hypothesize that the spatial and temporal relationship is important for skeleton action recognition, random sampling can be an overly difficult task in our scenarios.
For temporal colorization, the segment masking with a 15-frame segment length results in the best performance. 
For spatial colorization, the 10-joint masking performs the best. 
Therefore, we leverage these two masking strategies in the main experiments. 

Noted that we only have one style for person-level colorization, in which color information is not related to temporal and spatial information, we apply the random masking (25\%) for person stream in the main experiments.  

\begin{table*}[t]
\caption{Comparisons to state-of-the-art Supervised and Unsupervised Pretrain  skeleton action recognition methods on NTU RGB+D, NTU RGB+D 120, PKU-MMD, and NW-UCLA datasets.
}
\begin{center}
\begin{tabular}{|l|c|c|c|c|c|c|c|}
  \hline 
  \multirow{2}{*}{Method} &  \multicolumn{2}{c|}{ \textbf{NTU RGB+D}} & \multicolumn{2}{c|}{ \textbf{NTU RGB+D 120}} & \multicolumn{2}{c|}{ \textbf{PKU-MMD}}  & \multirow{2}{*}{\textbf{NW-UCLA}}\\
  \cline{2-7} & C-Subject & C-View & C-Subject & C-Setup & I & II &   \\
  \hline
  \hline
  \multicolumn{8}{|c|}{ \textbf{Supervised Method}} \\
  \hline
  \hline
  Actionlet ensemble \cite{wang2013learning} (TPAMI 2013) &--&--&--&--&--&--& 76.0 \\
  HBRNN-L \cite{du2015hierarchical} (CVPR 2015) & 59.1 & 64.0 &--&--&--&--& 78.5 \\
  Part-Aware LSTM \cite{Shahroudy_2016_CVPR} (CVPR 2016) & 62.9 & 70.3 & 25.5 & 26.3 & --& --&--\\
  ST-LSTM  \cite{liu2016spatio} (ECCV 2016)       & 69.2  & 77.7 & 55.7  & 57.9 &-- & --& --\\
  Ensemble TS-LSTM \cite{lee2017ensemble} (ICCV 2017) &  74.6 & 81.3 &--&--&--&--& 89.2 \\
  VA-RNN-Aug  \cite{zhang2019view} (TPAMI 2019) &79.8 & 88.9 &--&--&--&-- & 90.7 \\

  GCA-LSTM \cite{liu2017global} (CVPR 2017)         & 74.4  & 82.8 &-- &-- & --& --&-- \\
  ST-GCN \cite{yan2018spatial} (AAAI 2018)          & 81.5  & 88.3 & 70.7  & 73.2 &  84.1 & 48.2 &-- \\
  AS-GCN \cite{li2019actional} (CVPR 2019)            & 86.8  & 94.2 &-- & --& --&-- &-- \\
  2s-AGCN \cite{2sagcn2019cvpr} (CVPR 2019)          & 88.5 & 95.1 & 82.5 & 84.2 & 93.5 & 56.8 &-- \\
  2s-AGC-LSTM \cite{si2019attention} (CVPR 2019)     & 89.2  & 95.0 & --& -- &-- &-- & 93.3\\
  4s-Shift-GCN \cite{cheng2020skeleton} (CVPR 2020)    & 90.7  &  96.5 & 85.9  &  87.6 &-- & --&  94.6\\
  PA-ResGCN-B19 \cite{song2020stronger} (ACMMM 2020) & 90.9 &  96.0 & 87.3 &  88.3 &-- &-- &--\\
  MS-G3D \cite{liu2020disentangling} (CVPR 2020) & 91.5  & 96.2 & 86.9 & 88.4 & \textbf{95.0} & \textbf{59.2} &--\\
  CTR-GCN \cite{chen2021channel} (ICCV 2021) & 92.4  & 96.8 & 88.9 & 90.6 &-- &--  & 96.5\\
  Info-GCN \cite{chi2022infogcn} (CVPR 2022) & \textbf{93.0}  & \textbf{97.1} & \textbf{89.8} & \textbf{91.2} &-- &--  & \textbf{97.0}\\
  \hline
  \hline
  \multicolumn{8}{|c|}{ \textbf{Unsupervised Pretrain Method}} \\
  \hline
  \hline
  Li \textit{et al.} \cite{li2018unsupervised} (NeurIPS 2018) & 63.9 & 68.1 & --&-- & --& --& 62.5\\
  M$S^{2}$L \cite{lin2020ms2l} (ACMMM 2020) &    78.6     &   -- & --&  --& 85.2   & 45.7  &  86.8\\
  SKT \cite{zhang2022skeletal} (ICME 2022) & 83.1 & 91.2 & --&-- & --& --& --\\ 
  3s-CrosSCLR \cite{Li_2021_CVPR} (CVPR 2021) &  86.2   &  92.5&  80.5   &  80.4  & --&-- &--\\
  3s-AimCLR \cite{guo2022contrastive} (AAAI 2022)  &  86.9   &  92.8 &  80.1   & 80.9 & --&-- &-- \\
  2s-Colorization \cite{Yang_2021_ICCV} (ICCV 2021) & --  & -- & --&-- &-- &-- & 94.0\\
  3s-Colorization \cite{Yang_2021_ICCV} (ICCV 2021) & 88.0  & 94.9 & 78.5 & 79.3 & 91.5 & 54.0 & --\\
  \hline
   \textbf{2s-C-F Masked Colorization (Ours)} & --   & -- & --  &-- & --  &  -- &\textbf{95.0}\\
   \textbf{3s-C-F Masked Colorization (Ours)} & \textbf{89.1}   & \textbf{95.9} & \textbf{81.2}  & \textbf{82.4} & \textbf{93.3} & \textbf{57.7}  & --\\
  \hline
\end{tabular}
\end{center}

\label{tab:supervised result NTU}
\end{table*}

\subsection{Comparison with the State-of-the-art Methods}
We conduct extensive experiments under four settings, including unsupervised learning, semi-supervised learning, supervised learning, and transfer learning.  
We also study three skeleton cloud colorization configurations. 
`2s' means the combination of temporal and spatial steams, and `3s' stands for the combination of temporal, spatial, and person streams.

\subsubsection{Unsupervised Learning}
In the unsupervised setting, the feature extractor (\textit{i.e.,} the encoder $E(.)$) is trained with our proposed skeleton cloud colorization self-supervised representation learning approach. Then the feature representation is evaluated by the simple linear classifier $f(.)$, which is trained on the top of the frozen encoders $E(.)$. Such experimental setting for unsupervised learning has been widely adopted and practiced in prior studies \cite{kundu2019unsupervised, lin2020ms2l, nie2020view, zheng2018unsupervised}. Here for fair comparisons, we use the same setting as these prior works.

We compare our skeleton cloud colorization method with prior unsupervised methods on NTU RGB+D, NTU RGB+D 120, PKU-MMD, NW-UCLA, and UWA3D datasets, as shown in Table \ref{tab:unsupervised result NTU}. 
It can be seen that our proposed coarse-fine masked colorization method can achieve state-of-the-art performances on all five datasets. 
{PKU-MMD part II is a more challenging dataset with more skeleton noise caused by the view variation, and NTU RGB+D 120 is the largest dataset with multi-class.}
Our proposed method performs well on these two datasets, demonstrating the effectiveness of our proposed technique. 

\subsubsection{Semi-Supervised Learning}
We evaluate semi-supervised learning with the same protocol as in \cite{lin2020ms2l, si2020adversarial, li2020sparse, zhan2021spatial}, for fair comparisons. Under the semi-supervised setting, the encoder $E(.)$ is first pre-trained with colorized skeleton clouds and then jointly trained with the linear classifier $f(.)$ with a small ratio of action annotations. 
{We follow the pioneering semi-supervised skeleton work~\cite{si2020adversarial} to select the labeled training samples, which select specific percentages of training samples from each class to curate a labeled training set.}
Specifically, we derive labeled data by sampling $1\%$, $5\%$, $10\%$, $20\%$, $40\%$ data of each class from the training set of NTU RGB+D dataset, $1\%$, $5\%$, $10\%$, $20\%$, $40\%$, $50\%$ data of each class from the training set of NTU RGB+D 120 dataset, $1\%$, $10\%$ data of each class from the training set of PKU-MMD dataset, $1\%$, $5\%$, $10\%$, $15\%$, $30\%$, $40\%$ data of each class from the training set of NW-UCLA dataset, and $5\%$, $10\%$, $20\%$, $50\%$ data of each class from the training set of UWA-3D dataset, respectively.   

Table \ref{tab:semi-supervised result NTU}, Table \ref{tab:semi-supervised result NWUCLA}, Table \ref{tab:semi-supervised result NTU120}, Table \ref{tab:semi-supervised result UWA3D} and Table \ref{tab:semi-supervised result PKU} show experimental results for these five datasets, respectively.
It can be seen that most of the experimental results on all semi-setting of these five datasets achieve state-of-the-art performances. 
Even though the proposed framework does not outperform the state-of-the-art methods on four semi-settings (\textit{\textit{i.e.,} } semi-1 on NTU RGB+D 120 C-Subject, and semi-5 on NW-UCLA), we still can achieve competitive performance.  

{Additionally, for a much fairer comparison, we randomly select a specific percentage of training samples five times and subsequently perform five-fold experiments. 
These experiments involve our proposed methods and two state-of-the-art methods~\cite{Li_2021_CVPR, guo2022contrastive}, both of which provide the pre-trained models on the NTU RGB+D dataset. 
The experimental results can be found at the bottom of Table~\ref{tab:semi-supervised result NTU}, we can find that our proposed method consistently  outperforms~\cite{Li_2021_CVPR, guo2022contrastive} on the averaged results, further showing the effectiveness of our proposed methods. }

\begin{table}[t]
\begin{center}
\caption{
 Comparison of the transfer learning performance on PKUMMD part II dataset. (C-F stands for coarse-fine; `3s' means three-stream fusion)
}
\label{tab:transfer learning pkuv2}
\resizebox{0.48\textwidth}{!}{
\begin{tabular}{|l|c|c|}
  \hline 
  \multirow{2}{*}{Method} &  \multicolumn{2}{c|}{ \textbf{Transfer to PKU-MMD II}} \\
  \cline{2-3} & PKU-MMD I & NTU RGB+D 60 \\
  \hline
  LongT GAN \cite{zheng2018unsupervised} (AAAI 2018) & 43.6 & 44.8 \\
  M$S^{2}$L \cite{lin2020ms2l} (ACMMMM 2020) &    44.1      &  45.8    \\
  Skeleton-Contrastive \cite{thoker2021skeleton} (ACMMM 2020)  &  45.1     &  45.9  \\
  3s-Colorization \cite{Yang_2021_ICCV} (ICCV 2021) & 50.0    & 51.0  \\
  \hline
  \textbf{3s-C-F Masked Colorization (Ours)} & \textbf{58.0}    & \textbf{58.1}  \\
  \hline
\end{tabular}
}
\end{center}
\end{table}

\begin{table}[t]
\begin{center}
\caption{
 Comparison of the transfer learning performance on the Toyota Smarthome dataset.
(C-F stands for coarse-fine; `2s' means two-stream fusion)}
\label{tab:transfer learning toyota}
\begin{tabular}{|l|c|c|}
  \hline 
  \multirow{2}{*}{Methods} &  \multicolumn{2}{c|}{ \textbf{ Smarthome}} \\
  \cline{2-3} &   CS \   & CV1  \\  
  \hline
  UNIK~\cite{yang2021unik} (BMVC 2021) & 63.1    &  22.9 \\
  ViA~\cite{yang2022via} (arXiv 2022)   & 64.5    &  36.1 \\

  2s-Colorization [31] (ICCV 2021) & 70.4    & 42.1 \\
  
  \hline
  2s-C-F Masked Colorization (Ours) &  \textbf{72.6}    & \textbf{44.7}  \\
  
  \hline
\end{tabular}
\end{center}
\end{table}

\subsubsection{Supervised Learning}
Following the supervised evaluation protocol in \cite{lin2020ms2l, Li_2021_CVPR, guo2022contrastive}, we pre-train the encoder with our self-supervised masked skeleton colorization method and fine-tune the encoder and classifier by using labeled training data. Table \ref{tab:supervised result NTU} shows experimental results. 
We can find that the proposed coarse-fine masked skeleton colorization method outperforms the state-of-the-art self-supervised methods on all four datasets.
Though our framework is not designed for the supervised setting, its performance is comparable to state-of-the-art supervised skeleton action recognition methods on NTU RGB+D, PKU-MMD, and NW-UCLA datasets.

\subsubsection{Transfer Learning}
To further evaluate whether the proposed skeleton colorization method is able to gain knowledge of related tasks, we investigate the transfer learning performance of our model.

\noindent{\textbf{Transfer to PKU MMD II:}
Following the setting in \cite{lin2020ms2l}, we conduct the transfer learning experiments, which use NTU RGB+D and PKU-MMD I as the source datasets and PKU-MMD II as the target dataset.
Initially, we train the model using either the NTU RGB+D or PKU-MMD I Cross-Subject protocol. The pre-trained model is subsequently fine-tuned on the PKU-MMD II dataset.
Table~\ref{tab:transfer learning pkuv2} shows the transfer learning results. 
We can find that our proposed method achieves much better results than \cite{zheng2018unsupervised, lin2020ms2l, thoker2021skeleton}, showing that the feature extracted by our proposed method from a source dataset can improve classification accuracy on a different target set.}

\noindent{\textbf{Transfer to Smarthome~\cite{Das_2019_ICCV}:}
We also conduct the transfer learning experiments which use the NTU RGB+D as the source dataset and Smarthome as the target dataset. 
The experiments are conducted on the cross-subject (CS) and cross-view1 (CV1) evaluation protocols.
The experimental results are shown in Table~\ref{tab:transfer learning toyota}.
It can be seen that our proposed method achieves state-of-the-art performance, demonstrating the effectiveness of our proposed method when transferred to related datasets. }

{It should be noted that the Toyota Smarthome skeleton data consists of only 13 skeleton joints. To ensure the input data from Toyota Smarthome matches the size of the NTU RGB+D dataset, we applied interpolation and zero padding.}

\section{Conclusion}
\label{sec: conclusion}
In this paper, we address self-supervised representation learning in skeleton action recognition and design a novel coarse-fine masked skeleton cloud colorization method that is capable of learning skeleton representations from unlabeled data.
Specifically,  we obtain colored skeleton cloud representations by stacking skeleton sequences to 3D skeleton clouds and colorizing each point according to its temporal and spatial orders in the skeleton sequences.
Besides, a two-steam pretrained framework that leverages coarse-grained colorization and fine-grained colorization is also introduced to learn multi-scale spatial-temporal features.
Additionally, we introduce a Masked Skeleton Cloud Repainting task to pretrain the designed auto-encoder framework.
Extensive experiments over different datasets demonstrate that our proposed method achieves superior unsupervised and semi-supervised action recognition performance.

\ifCLASSOPTIONcaptionsoff
  \newpage
\fi

{
\bibliographystyle{IEEEtran}
\bibliography{egbib}
}

\begin{IEEEbiography}[{\includegraphics[width=1in,height=1.25in,clip,keepaspectratio]{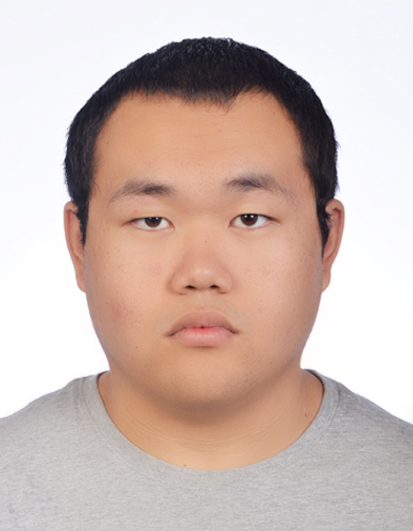}}]{Siyuan Yang} received the BEng degree from Harbin Institute of Technology and the MSc degree from Nanyang Technological University. He is currently pursuing the Ph.D. degree with the Interdisciplinary Graduate Programme, Nanyang Technological University. His research interests include computer vision, action recognition, and human pose estimation.  
\end{IEEEbiography}

\begin{IEEEbiography}[{\includegraphics[width=1in,height=1.25in,clip,keepaspectratio]{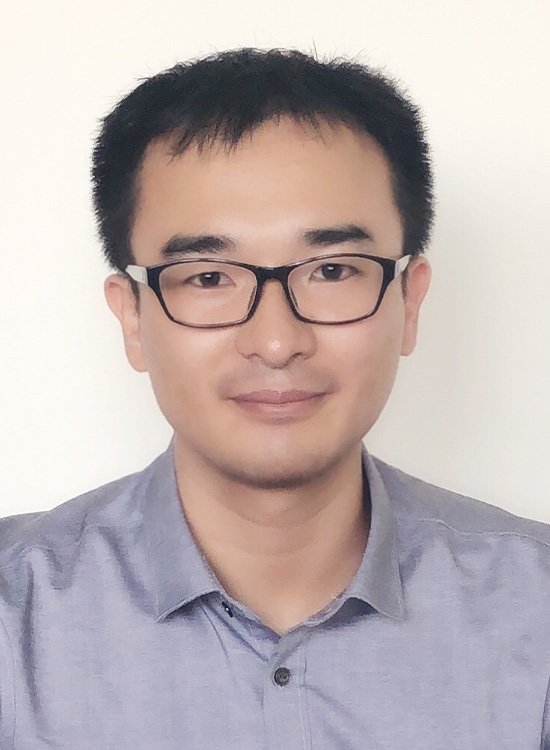}}]{Dr. Jun Liu} is an Assistant Professor with Singapore University of Technology and Design. He received the PhD degree from Nanyang Technological University, the MSC degree from Fudan University, and the BEng degree from Central South University. His research interests include computer vision and artificial intelligence. He is an Associate Editor of IEEE Transactions on Image Processing, and area chair of ICLR, ICML, NeurIPS, and WACV in 2022 and 2023.\end{IEEEbiography}

\begin{IEEEbiography}[{\includegraphics[width=1in,height=1.25in,clip,keepaspectratio]{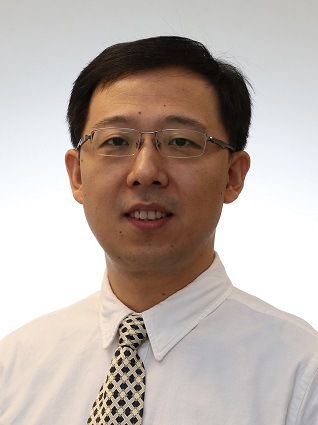}}]{Dr. Shijian Lu} is an Associate Professor in the School of Computer Science and Engineering, Nanyang Technological University in Singapore. He obtained the PhD from the Electrical and Computer Engineering Department, National University of Singapore. Dr Shijian Lu’s major research interests include image and video analytics, visual intelligence, and machine learning. He has published more than 100 international refereed journal and conference papers and co-authored over 10 patents in these research areas. Hi was also a top winner of the ICFHR2014 Competition on Word Recognition from Historical Documents, ICDAR 2013 Robust Reading Competition, etc. Dr Lu is an Associate Editor for the journal Pattern Recognition (PR) and Neurocomputing. He has also served in the program committee of several conferences, e.g., the Senior Program Committee of the International Joint Conferences on Artificial Intelligence (IJCAI) 2018 – 2021, etc.\end{IEEEbiography}

\begin{IEEEbiography}[{\includegraphics[width=1in,height=1.25in,clip,keepaspectratio,trim=0cm 0cm 0cm 0cm]{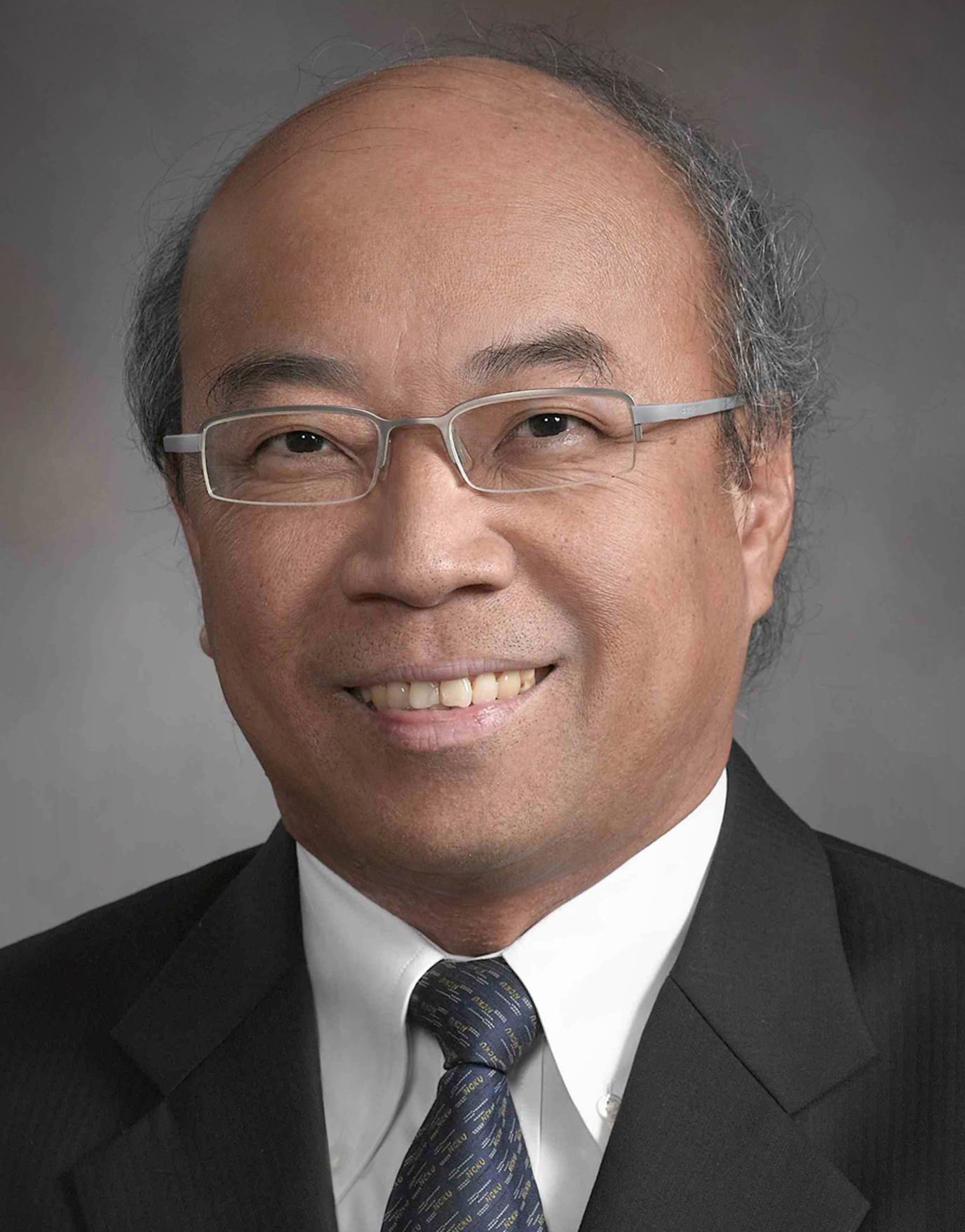}}]{Prof. Meng Hwa Er} received the B.Eng. degree (with first class Hons.) in electrical from the National University of Singapore (NUS), Singapore and the Ph.D. degree in electrical and computer engineering from the University of Newcastle (UON), Callaghan, NSW, Australia, under a UON Postgraduate Research Scholarship. He has a long track record of distinguished academic and administrative services to Nanyang Technological University (NTU), Singapore, in various top management positions, including the Deputy President, Acting Provost, Vice President (International Affairs), the Dean of School of Electrical and Electronic Engineering and the Acting Dean of Graduate Studies, just to name a few. He had also founded the Centre for Signal Processing funded by NSTB and NTU Satellite Engineering Programme and was the founding Dean of the College of Engineering. He is currently a Professor of electrical and electronic engineering and the Director of Centre for Information Sciences and Systems (CISS), the School of Electrical and Electronic Engineering. He is a Member of the Advisory Board, NTU Academic Council.

His outstanding contribution to engineering, education and public service is underpinned by an impressive research track record in the fields of smart antenna array signal processing, satellite communications, computer vision and optimisation techniques. He has secured and completed a total of 32 research projects with a cumulative grant value of approximately S\$49 million and has authored or co-authored more than 266 papers in international journals and conference proceedings and six book chapters. He also holds five patents. He spearheaded the development of core capabilities in satellite communication technology and Low Earth Orbit (LEO) micro-satellite technology in collaboration with DSO National Labs which led to the successful launch of the first Singapore’s design and build experimental micro-satellite (X-Sat) in 2011. It has put NTU and Singapore on the world map in micro-satellite technology. His contribution in micro-satellite technology has been featured in NTU's 30th anniversary celebration commemorative book - 30 Years of Momentum.

He is a Life Fellow of IEEE and IES, an Honorary Fellow of IEE and ASEAN Federations of Engineering Organisations (AFEO), a Fellow of Academy of Engineering, Singapore (SAEng) and ASEAN Academy of Engineering and Technology (AAET), a registered Professional Engineer (PE) and ASEAN Chartered Professional Engineer (ACPE).
\end{IEEEbiography}

\begin{IEEEbiography}[{\includegraphics[width=1in,height=1.25in,clip,keepaspectratio]{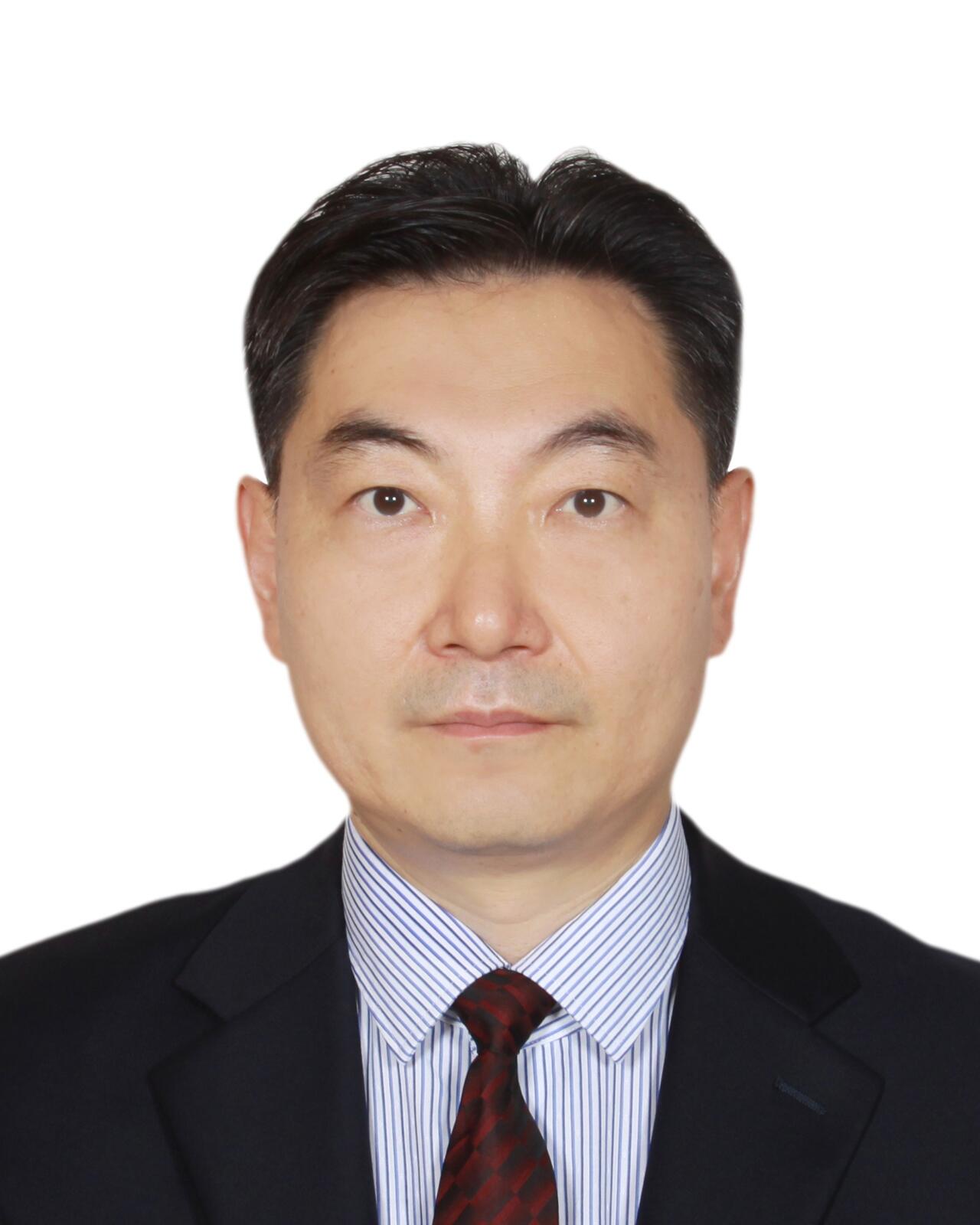}}]
{Prof. Yongjian Hu} received the Ph.D. degree in communication and information systems from South China University of Technology in 2002. Between 2000 and 2004, he visited City University of Hong Kong four times as a researcher. From 2005 to 2006, he worked as Research Professor in SungKyunKwan University, South Korea. From 2006 to 2008, he worked as Research Professor in Korea Advanced Institute of Science and Technology, South Korea. From 2011 to 2013, he worked as Marie Curie Fellow in the University of Warwick, UK. Now He is full Professor with the School of Electronic and Information Engineering, South China University of Technology, China. He is also a research scientist with China-Singapore International Joint Research Institute. Dr. Hu is Senior Member of IEEE and has published more than 130 peer reviewed papers. His research interests include image forensics, information security, and deep learning.

\end{IEEEbiography}

\begin{IEEEbiography}[{\includegraphics[width=1in,height=1.25in,clip,keepaspectratio]{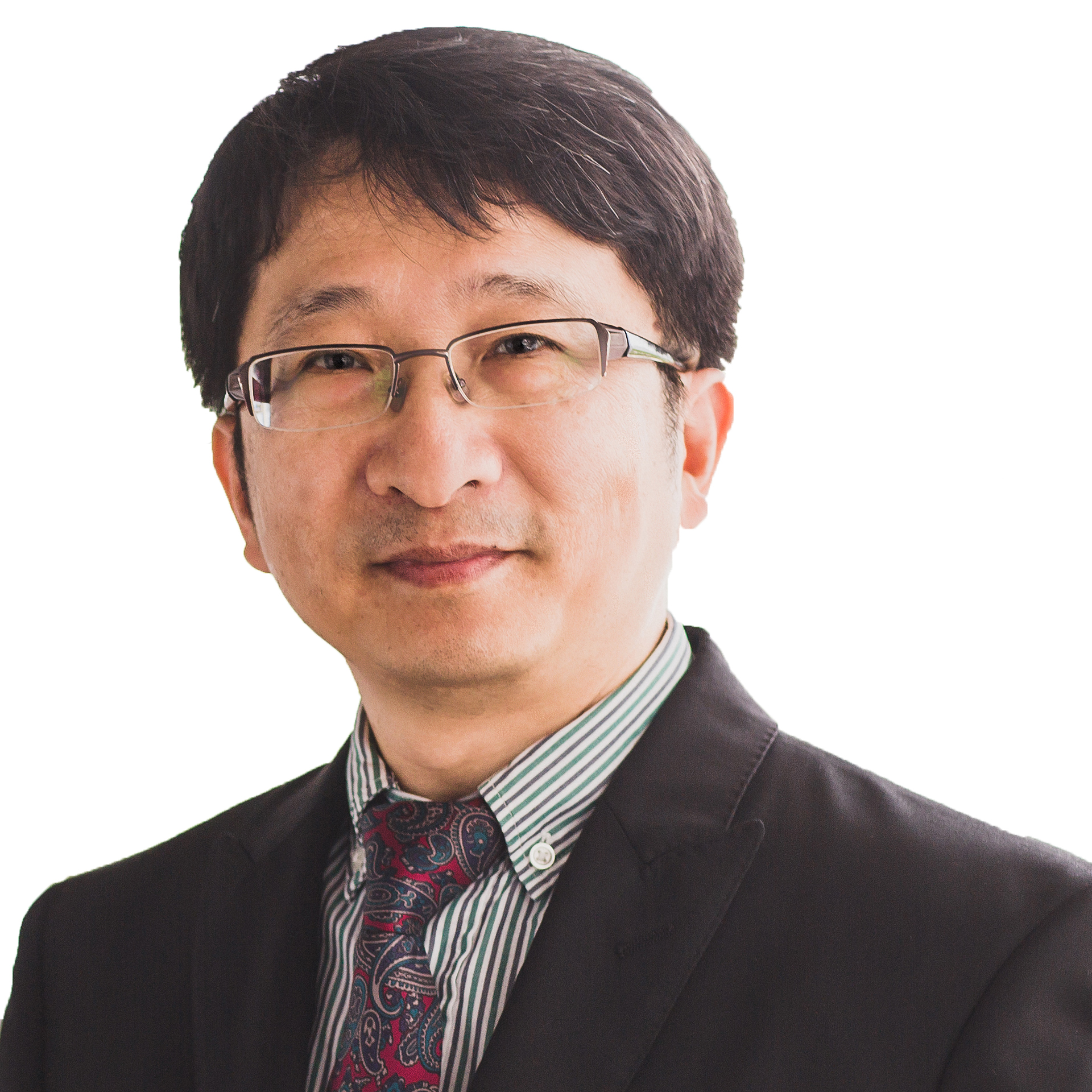}}]
{Prof. Alex Kot} has been with the Nanyang Technological University, Singapore since 1991. He was Head of the Division of Information Engineering and Vice Dean Research at the School of Electrical and Electronic Engineering. Subsequently, he served as Associate Dean for College of Engineering for eight years. He is currently Professor and Director of Rapid-Rich Object SEarch (ROSE) Lab and NTU-PKU Joint Research Institute. He has published extensively in the areas of signal processing, biometrics, image forensics and security, and computer vision and machine learning.

Dr. Kot served as Associate Editor for more than ten journals, mostly for IEEE transactions. He served the IEEE SP Society in various capacities such as the General Co-Chair for the 2004 IEEE International Conference on Image Processing and the Vice-President for the IEEE Signal Processing Society. He received the Best Teacher of the Year Award and is a co-author for several Best Paper Awards including ICPR, IEEE WIFS and IWDW, CVPR Precognition Workshop and VCIP. He was elected as the IEEE Distinguished Lecturer for the Signal Processing Society and the Circuits and Systems Society. He is a Fellow of IEEE, and a Fellow of Academy of Engineering, Singapore 

\end{IEEEbiography}

\begin{figure*}[h]
\centering
\includegraphics[trim=1.5cm 1cm 0cm 1cm,clip,width=0.95\textwidth]{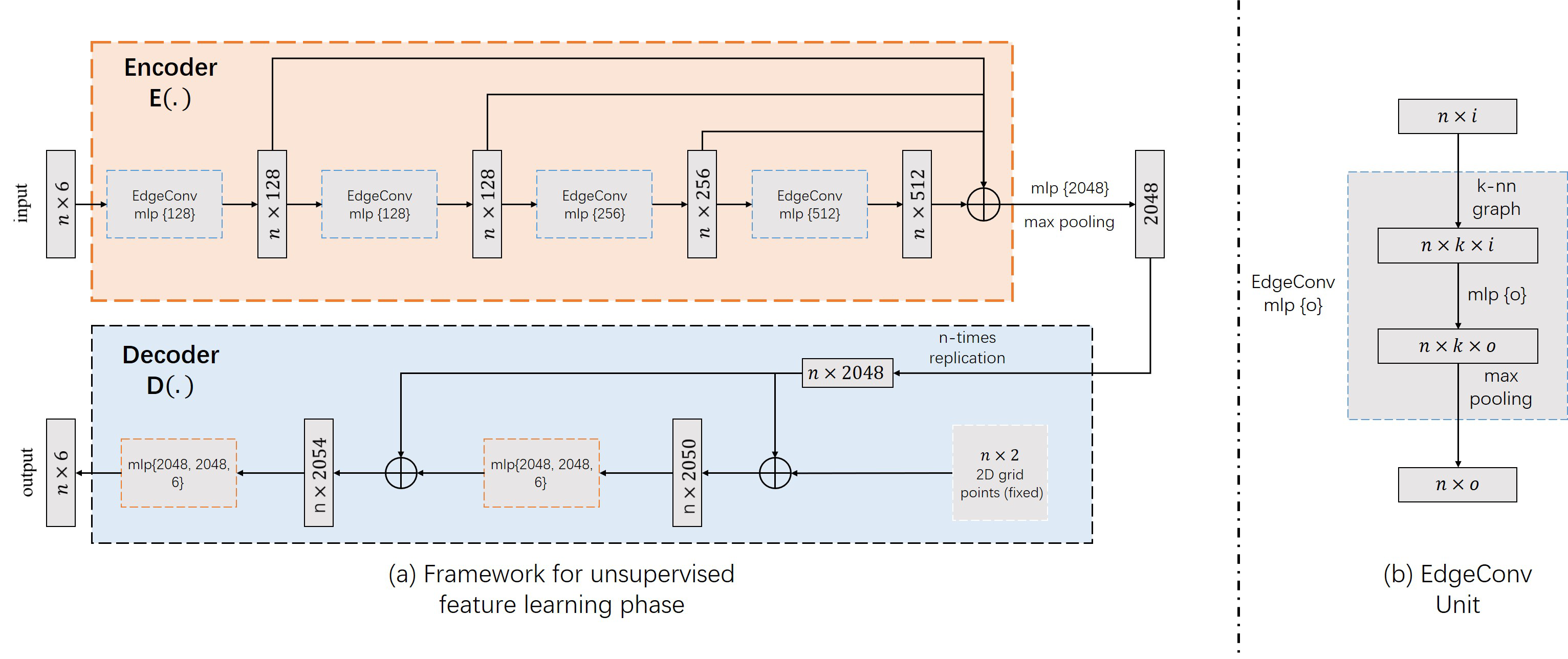}
  \caption{Details of our network architectures. (a) shows the framework of self-supervised feature learning, where the input is the raw skeleton cloud or partially colorized skeleton cloud, and the output is the repainted skeleton cloud. ($n$ is the number of points)
 (b) shows details of the EdgeConv Unit that is used in (a). The input of EdgeConv is a tensor with shape $n\times i$, and the output is a tensor with shape $n\times o$. ($n$ is the number of points, $i$ and $o$ denote the dimensions of input and output features per point, respectively)
  }
\label{fig:Unsupervised feature learning phase framework}
\end{figure*}

\begin{figure*}[h]
\begin{center}
\includegraphics[trim=1.52cm 0cm 0cm 0cm,clip,width=0.88\textwidth]{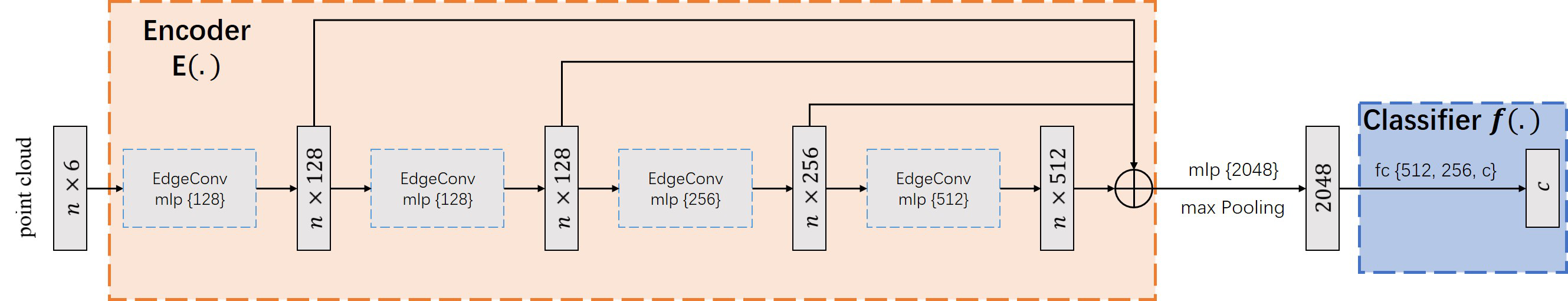}
\end{center}
  \caption{The framework of linear classifier training. The encoder is pre-trained in the self-supervised feature learning phase, as illustrated in Fig. \ref{fig:Unsupervised feature learning phase framework}. The encoder is then frozen under the unsupervised setting, and it is fine-tuned together with the linear classifier under the semi-supervised, supervised, and transfer learning settings. ($n$ is the number of points, c is the number of action categories)
  }
\label{fig:finetune phase framework}
\end{figure*}

\newpage
\appendices
\section{Detailed Network Architecture}

As described in Section~\ref{sec: implentation} of the main manuscript, our network training consists of two phases including an self-supervised feature learning phase and a linear classifier training phase. 
For self-supervised feature learning, we use DGGNN \cite{wang2019dynamic} as our encoder $E(.)$, and use the decoder of FoldingNet \cite{yang2018foldingnet} as our decoder $D(.)$.
The inputs of DGCNN and FoldingNet are $N \times 3$ matrices that are composed of 3D positions $(x, y, z)$. While the inputs to our network consist of both 3D positions $(x, y, z)$ and color information $(r, g, b)$ with a $N \times 6$ matrix, we double the size of each network layer.

Following previous works on self-supervised action recognition \cite{kundu2019unsupervised, lin2020ms2l, nie2020view}, we place a linear classifier $f(.)$ (with 3 FC layers) on top of the encoder to perform action recognition. Fig. \ref{fig:finetune phase framework} shows the network architecture. 

Under the unsupervised setting, 
after the encoder has been trained via self-supervised feature learning (shown in Fig. \ref{fig:Unsupervised feature learning phase framework}),
we then train the linear classifier only, and keep the encoder frozen (shown in Fig. \ref{fig:finetune phase framework}). 
Under the semi-supervised, supervised and transfer learning settings, the encoder is first trained via self-supervised feature learning and then fine-tuned together with the linear classifier.

\section{More ablation studies on NW-UCLA}
Given the substantial number of tables in the manuscript, we have opted to relocate the ablation studies on NW-UCLA to the appendix.

\subsection{Effectiveness of Our Skeleton Colorization:}
Table~\ref{tab:Abla semi NWUCLA}  and Table~\ref{tab: abla_unsup_and_sup ucla} show experimental results on the NW-UCLA dataset.
Both temporal-level and spatial-level colorization strategies demonstrate a notable performance enhancement compared to the baseline on all the unsupervised, semi-supervised, and supervised learning settings.
Moreover, when compared with the two stronger baselines (i.e., motion prediction and masked autoencoder), our proposed color repainting strategy consistently surpasses them by a larger margin.

\begin{table*}[t]
\begin{center}
\caption{Comparisons of different network configurations' results with the semi-supervised setting on NW-UCLA dataset. (`\textit{TS}': Temporal Stream; `\textit{SS}': Spatial Stream; `2s' means two-stream fusion; $v./c.$ denotes the number of labeled videos per class)
}
\label{tab:Abla semi NWUCLA}
\resizebox{0.95\textwidth}{!}{
\begin{tabular}{l|c|c|c|c|c|c}
  \hline 
  Method &  $1\% \; (1 \;v_{.}/c_{.})$ &  $5\% \; (5 \;v_{.}/c_{.})$ &  $10\% \; (10 \;v_{.}/c_{.})$ &  $15\% \; (15 \;v_{.}/c_{.})$ &  $30\% \; (30 \;v_{.}/c_{.})$ &  $40\% \; (40 \;v_{.}/c_{.})$ \\
  \hline
  Baesline-Semi  & 34.3   &  46.4  &  54.9 &  61.8 & 69.1 & 70.2 \\
    Motion prediction  & 38.8   &  51.6  &  58.5 &  64.6 & 
  69.7 & 73.0 \\
  Mask autoencoder  & 37.8   &  49.7  &  55.3 &  64.6 & 71.1 & 73.8 \\
  \hline
  `\textit{TS}' Colorization             & 40.6  &  55.9 &  71.3 &  74.3  & 81.4 & 83.6 \\
  `\textit{SS}' Colorization   & 39.1  &  54.2 &  66.3 &  70.2 & 79.1 & 80.8 \\
  \hline
  2s-Colorization  &  \textbf{41.9}  &    \textbf{57.2}    &    \textbf{75.0}    &    \textbf{76.0}    &  \textbf{83.0}    &   \textbf{84.9} \\
  \hline
\end{tabular}
}
\end{center}
\end{table*}

\begin{table}[t]
\begin{center}
\caption{Comparisons of different network configurations' results with unsupervised and supervised settings on NW-UCLA dataset. (`\textit{TS}': Temporal Stream; `\textit{SS}': Spatial Stream; `2s' means two-stream fusion)
}
\label{tab: abla_unsup_and_sup ucla}
\begin{tabular}{|l|c|} 
  \hline 
  Dataset  &  NW-UCLA \\
  \hline
  \multicolumn{2}{|c|}{ \textbf{Unsupervised Setting}} \\
  \hline
  Baseline-U      &  78.6      \\
  Motion Prediction-U        & 82.1 \\
   Masked Autoencoder-U       & 83.6 \\
  \hline
  `\textit{TS}' Colorization    &     90.1   \\
  `\textit{SS}' Colorization      &     87.0      \\
  \hline
  2s-Colorization & \textbf{91.1}  \\
  \hline
  \multicolumn{2}{|c|}{ \textbf{Supervised Setting}} \\
  \hline
  Baseline-S       &  83.8        \\
     Motion Prediction-S       & 84.1 \\
   Masked Autoencoder-S     &    86.6\\
  \hline
  `\textit{TS}' Colorization &     92.7      \\
  `\textit{SS}' Colorization      &     90.4       \\
  \hline
    2s-Colorization & \textbf{94.0} \\
  \hline
\end{tabular}
\end{center}
\end{table}

\subsection{Selection of Segment Size and Body Part Scale:}
As shown in Table~\ref{tab: Segment Body 2}, on the NW-UCLA dataset, we can achieve the same conclusion as in the main paper that 5-frame coarse-grained temporal colorization and 10-body-part (scale 1) coarse-grained spatial colorization achieve the best performance, respectively.

\begin{table}[t]
\caption{
Ablation studies on the Segment Size and Body Part Scale. The experiments conduct on the UW-UCLA dataset under the unsupervised setting.
}
\label{tab: Segment Body 2}
\begin{center}
\begin{tabular}{|cc|cc|}
  \hline 
  Segment Step Size &  NW-UCLA & Spatial scale & NW-UCLA\\
  
  \hline
  2   & 90.1 & 1 (10 parts) & \textbf{87.9} \\
  4   & 89.2  & 2 (6 parts)  & 86.8\\
  5  & \textbf{91.0} & &\\
  8  & 90.8  & &\\
  10 & 90.3  & &\\
  20 & 89.9  & &\\
  \hline
\end{tabular}
\end{center}
\end{table}

\subsection{Effectiveness of Masking Strategy and Coarse-Fine Alignment framework:}
As shown in Table~\ref{tab: Cosine-to-Fine 2}, the masking task enhances the performance for both two colorization streams on the NW-UCLA dataset. 
Additionally, the proposed Coarse-Fine Alignment framework contributes to further improvements.

\begin{table}[t]
\begin{center}
\caption{Linear evaluation results compared with Skeleton Colorization~\cite{Yang_2021_ICCV} on NW-UCLA. `$\Delta$' represents the gain compared to~\cite{Yang_2021_ICCV} with the same stream data. (C-F stands for coarse-fine; `\textit{TS}':Temporal Stream; `\textit{SS}':Spatial Stream; `2s' means three-stream fusion)
}
\label{tab: Cosine-to-Fine 2}
\begin{tabular}{|l|cc|}
  \hline 
  \multirow{1}{*}{Method} & \multicolumn{2}{c|}{\textbf{NW-UCLA}} \\
  \cline{2-3} 
  & Acc. & $\Delta$ \\
  \hline
  `\textit{TS}' Colorization \cite{Yang_2021_ICCV}   & 90.1  & \\
  `\textit{TS}' Masked Colorization    & 90.5 &  $\uparrow$ 0.4   \\
  `\textit{TS}' C-F Masked Colorization  (Ours)  & 91.0 & $\uparrow$ 0.9  \\
  \hline
  `\textit{SS}' Colorization \cite{Yang_2021_ICCV}  &  87.0 &   \\
  `\textit{SS}' Masked Colorization   & 87.3 & $\uparrow$ 0.3   \\
  `\textit{SS}' C-F Masked Colorization  (Ours)     & 87.9  & $\uparrow$ 0.9      \\
  \hline
  2s-Colorization \cite{Yang_2021_ICCV} &  91.1& \\
  2s-Masked Colorization  & 91.4 &  $\uparrow$ 0.3  \\
  \textbf{2s-C-F Masked Colorization (Ours)}  & \textbf{92.0} &  $\uparrow$ 0.9 \\
  \hline
\end{tabular}
\end{center}
\end{table}

\subsection{Effectiveness of Different Masking Strategies and
Masking Ratios:}
On the NW-UCLA dataset, segment masking with a 15-frame length and 15-frame masking both produce optimal results for temporal colorization. 
For spatial colorization, both 10-joint masking and 6-part masking yield equally best performances.
For consistency with the NTU RGB+D dataset, we leverage these 15-frame segment masking and 10-joint masking strategies in the main experiments. 

\begin{table}[t]
\begin{center}
\caption{Ablation study on temporal masking strategy, including the temporal random masking (Fig.~\ref{fig:masking} (b)), frame-only masking (Fig.~\ref{fig:masking} (c)), and segment masking (Fig.~\ref{fig:masking} (d)). The experiments conduct on NW-UCLA under the unsupervised setting.
}
\label{tab:temporal masking ucla}
\resizebox{0.49\textwidth}{!}{
\setlength\tabcolsep{2pt}
\begin{tabular}{|c c|c c|c c|}
  \hline 
\multicolumn{2}{|c|}{ \textbf{(b) Random Masking}} & \multicolumn{2}{c|}{ \textbf{(c) Frame-only Masking}} & \multicolumn{2}{c|}{ \textbf{(d) Segment Masking}} \\
  \hline
    \hline
    Mask Ratio & NW-UCLA  & Mask Frame Number & NW-UCLA  & Maks Segment Length & NW-UCLA \\
  \hline

    0.25 & 91.0 &  5 & 89.5 &  5 & 89.7 \\
    0.50 & 89.5  & 10 & 89.5 & 10 & 90.1\\
    0.75  & 90.1  & 15 & \textbf{91.2}   & 15 & \textbf{91.2} \\
         &     &  20 & 90.1   & 20 & 89.9  \\
         &     &  30 & 89.7 & 30 & 89.5\\
  \hline
\end{tabular}
}
\end{center}
\end{table}

\begin{table}[h]
\begin{center}
\caption{Ablation study on spatial masking strategy, including the spatial random masking (Fig.~\ref{fig:masking} (f)), joint-only masking (Fig.~\ref{fig:masking} (g)), and body-part masking (Fig.~\ref{fig:masking} (h)). The experiments conduct on NW-UCLA under the unsupervised setting.
}
\label{tab:spatial masking 2}
\resizebox{0.49\textwidth}{!}{
\setlength\tabcolsep{2pt}
\begin{tabular}{|c c|c c|c c|}
  \hline 
\multicolumn{2}{|c|}{ \textbf{(f) Random Masking}} & \multicolumn{2}{c|}{ \textbf{(g) Joint-only Masking}} & \multicolumn{2}{c|}{ \textbf{(h) Body-part Masking}} \\
  \hline
  \hline 
    Mask Ratio & NW-UCLA  & Mask Joint Number  & NW-UCLA  & Mask Part Number & NW-UCLA \\
  \hline
   0.25  & 87.9   & 5  & 87.9  & 2 & 87.5 \\
   0.50  & 87.4   & 10   & \textbf{88.3} & 4  & 87.7\\
   0.75  & 86.2   & 15 & 87.5  & 6  & \textbf{88.3}\\
      &      &   &    & 8   & 87.1\\
  \hline
\end{tabular}
}
\end{center}
\end{table}

\section{Experiments with different multi-colorization strategies}
We've explored four distinct strategies to consolidate the spatial-temporal or the spatial-temporal-person colorization into a single skeleton cloud.

\noindent (1) `TS-S'-mul: 
In this strategy, we create a spatial-temporal colorized skeleton cloud by multiplying the spatial and temporal colors of each joint.

\noindent (2) `TSP-S'-spe: We allocate unique colors to all points, from the order of [1, ..., TJP].

\noindent (3) `TS-S'-concat: We concatenate the temporal and spatial colors together to leverage a 6-items colorization code to stand for the temporal and spatial order information.

\noindent (4) `TSP-S'-concat: We concatenate the temporal, spatial, and person-level colors, forming a 9-items colorization code that represents the temporal, spatial, and person order information. 

The experimental results in Table~\ref{tab: colorization together} show that combined strategies don't deliver superior results compared to single-stream colorization methods (\textit{`T-Stream' (TS)} and \textit{`S-Stream' (SS)}).
We hypothesize that this might stem from the challenges the neural network encounters when managing a surfeit of diverse color data or an overload of information related to each skeleton point during the self-supervised learning phase.
All these experiments are conducted with random masking (25\%) on the NTU RGB+D dataset. 

\begin{table}[h]
\begin{center}
\caption{Comparison with different multi-colorization strategies on the NTU RGB+D dataset. (`\textit{TS}': Temporal Stream; `\textit{SS}': Spatial Stream)
}
\label{tab: colorization together}
\begin{tabular}{|l|c|c|} 
  \hline 
    Method & NTU-CS  &  NTU-CV  \\
  \hline

   `TS-S'-mul Colorization       & 70.1    &  79.8     \\
   `TS-S'-concat Colorization    & 70.2    &  79.4     \\
   `TSP-S'-spe Colorization        & 68.6    &  78.9     \\
   `TSP-S'-concat Colorization    & 70.5    &  79.9     \\
   
  \hline
   `\textit{TS}' Colorization             & \textbf{72.1}    &  \textbf{81.1}     \\
   `\textit{SS}' Colorization            & 69.6    &  80.1     \\
  \hline
\end{tabular}
\end{center}
\end{table}

\section{Experiments with different colorization schemes  and different orders of RGB values}
We conduct the experiment with the other colorization scheme $r = t/T, g = t/T, b = t/T$, and the experimental results are shown in Table~\ref{tab: different colorization schemes 2}.
It can be seen that our proposed colorization achieves better performance on both the temporal colorization and spatial colorization streams. 
In the colorization scheme $r = t/T, g = t/T, b = t/T$, the color distribution is from blue to white, thus containing less order information. 

Furthermore, we experimented with varying the order of RGB values, specifically using the sequences $r,b,g$ and $b,g,r$. The experimental results are shown in Table~\ref {tab: different colorization schemes 2}. We can find that all these three orders achieve similar performance. 
We hypothesize that the linear progression inherent in these color sequences facilitates order learning.
For an easier understanding, we will still keep the conventional order $r,b,g$. 
All these experiments are conducted with random masking (25\%) on the NTU RGB+D dataset. 

\begin{table}[h]
\begin{center}
\caption{Comparison with different colorization schemes and different orders of RGB values on the NTU RGB+D dataset. 
}
\label{tab: different colorization schemes 2}
\resizebox{0.49\textwidth}{!}{
\begin{tabular}{|l|c|c|c|c|c|} 
\hline
\multirow{2}{*}{Color Scheme} & \multirow{2}{*}{Color Order} & \multicolumn{2}{c|}{ \textbf{Temporal colorization}} & \multicolumn{2}{c|}{ \textbf{Spatial colorization}} \\
  \cline{3-6} 
   &   & NTU-CS  &  NTU-CV  & NTU-CS  &  NTU-CV \\
  \hline
   $r,g,b = t/T$ & [r,g,b]   & 68.5   &  78.3      &  66.9   &  77.1         \\
   \hline
   Ours   & [r,b,g]     & 71.1   &  80.8      &  69.5    &   80.0     \\
   Ours    & [b,g,r]    & 71.2   &  80.4    & \textbf{70.3}    &   \textbf{80.7}   \\
   \hline
   Ours      & [r,g,b]    & \textbf{72.1}   &  \textbf{81.1}    &  69.6   &  80.1       \\
  \hline
\end{tabular}
}
\end{center}
\end{table}

\section{proof of the ability to learn the temporal dependency}
In order to prove our method's ability to learn the temporal dependency, we first conduct the experiment with the reverse-order sequence. 
Utilizing a pre-trained model developed with the standard color order skeleton cloud, we conducted experiments in the unsupervised linear evaluation setting.
As the point cloud data is unordered data, so during the experiment, we maintain the original $[x, y, z]$ values while reversing the order for temporal-based color values. 
The experimental results are shown in Table~\ref{tab: random and reverse}.
It is a fact that the reversed color order still provides temporal information, so our proposed method still can achieve relatively strong performance when dealing with the skeleton cloud with the revised temporal color information.
Additionally, to further explain the ability to learn temporal dependency, we also conduct the experiment with the random-order temporal color information. 
It can be seen that, with the random colored information, we can only achieve much lower performance than the normal one and the reverse one. 
All the above two observations show that our proposed self-supervised method is able to learn the temporal dependency.
All these experiments are conducted with random masking (25\%) on the NTU RGB+D dataset. 

\begin{table}[h]
\begin{center}
\caption{Comparison of the skeleton cloud with random/reverse temporal color on the NTU RGB+D dataset. (C-F stands for coarse-fine; `\textit{TS}': Temporal Stream; `\textit{SS}': Spatial Stream)
}
\label{tab: random and reverse}
\begin{tabular}{|l|c|c|} 
  \hline 
    Method & NTU-CS  &  NTU-CV  \\
  \hline
   `\textit{TS}'-random Colorization    & 65.6  & 75.1      \\
   `\textit{TS}'-reverse Colorization    & 70.7  & 79.9             \\
   `\textit{TS}' Colorization             & \textbf{72.1}  & \textbf{81.1}    \\
     \hline
   `C-F-\textit{TS}'-random Colorization    & 67.6 &  76.3   \\  
   `C-F-\textit{TS}'-reverse Colorization   & 72.0 &  81.3  \\
   `C-F-\textit{TS}' Colorization            & \textbf{73.2}   & \textbf{82.6}   \\
  \hline

\end{tabular}
\end{center}
\end{table}

\section{Further details on the formulation of the colorization}
\textbf{Spatial Colorization.}
For spatial colorization, the distribution of the values of R, G, and B channels can be calculated as follows:
\begin{equation}
r^{s}_{t, j} = \left\{
\begin{aligned}
-2\times (j/J) + 1 & ,\;\mbox{if $j <= J/2$}\\
0  & ,\;\mbox{if $j > J/2$}\\
\end{aligned}
\right. 
\end{equation}
\begin{equation}
g^{s}_{t, j} = \left\{
\begin{aligned}
2\times (j/J) & ,\;\mbox{if $j <= J/2$}\\
-2\times (j/J) + 2  & ,\;\mbox{if $j > J/2$}\\
\end{aligned}
\right.
\end{equation}
\begin{equation}
b^{s}_{t, j} = \left\{
\begin{aligned}
0 & ,\;\mbox{if $j <= J/2$}\\
2\times (j/J) - 1  & ,\;\mbox{if $j > J/2$}\\
\end{aligned}
\right.
\end{equation}
With the increase of the spatial order index of the joint in the skeleton, points will be assigned with different colors that change from red to blue and to green gradually, which is able to represent the linear order information for the joint spatial order and facilitate the learning of spatial order information.

\noindent\textbf{Spatial Coarse-Grained Colorization.}
The value distributions of R, G, and B channels can be formulated as follows:

\begin{equation}
r^{cs}_{t, \mathcal{p}} = \left\{
\begin{aligned}
-2\times (\mathcal{p}/\mathcal{P}) + 1 & ,\;\mbox{if $\mathcal{p} <= \mathcal{P}/2$}\\
0  & ,\;\mbox{if $\mathcal{p} > \mathcal{P}/2$}\\
\end{aligned}
\right.
\end{equation}

\begin{equation}
g^{cs}_{t, \mathcal{p}} = \left\{
\begin{aligned}
2\times (\mathcal{p}/\mathcal{P}) & ,\;\mbox{if $\mathcal{p} <= \mathcal{P}/2$}\\
-2\times (\mathcal{p}/\mathcal{P}) + 2  & ,\;\mbox{if $\mathcal{p} > \mathcal{P}/2$}\\
\end{aligned}
\right.
\end{equation}

\begin{equation}
b^{cs}_{t, \mathcal{p}} = \left\{
\begin{aligned}
0 & ,\;\mbox{if $\mathcal{p} <= \mathcal{P}/2$}\\
2\times (\mathcal{p}/\mathcal{P}) - 1  & ,\;\mbox{if $\mathcal{p} > \mathcal{P}/2$}\\
\end{aligned}
\right.
\end{equation}
\noindent With this colorization scheme, we can assign different colors to points from different body parts based on the body part index $\mathcal{P}$, as illustrated in Fig. \ref{fig:color_pipeline_spatial} (b).

\noindent\textbf{Temporal Coarse-Grained Colorization.}
We employ the colorization scheme (Fig.~\ref{fig:colorize} (a)) to colorize segment-level information as follow:

\begin{equation}
r^{ct}_{s, j} = \left\{
\begin{aligned}
-2\times (s/S) + 1 &, \;\mbox{if $s <= S/2$}\\
0  &, \;\mbox{if $s > S/2$}\\
\end{aligned}
\right.
\end{equation}

\begin{equation}
g^{ct}_{s, j} = \left\{
\begin{aligned}
2\times (s/S) & ,\;\mbox{if $s <= S/2$}\\
-2\times (s/S) + 2  &, \;\mbox{if $s > S/2$}\\
\end{aligned}
\right.
\end{equation}

\begin{equation}
b^{ct}_{s, j} = \left\{
\begin{aligned}
0 & ,\;\mbox{if $s <= S/2$}\\
2\times (s/S) - 1  & ,\;\mbox{if $s > S/2$}\\
\end{aligned}
\right.
\end{equation}
Here, we use the $P_{\tau c}$ to stand for the temporal coarse-grained colorized skeleton cloud. 
Temporal Coarse-Grained Colorization allocates distinct colors to points from different segments, as shown in Fig.~\ref{fig:color_pipeline_temporal} (b).
Consequently, the proposed temporal coarse-grained colorized skeleton cloud $P_{\tau c}$ effectively captures the temporal dependency information at the segment level. 

\end{document}